\documentclass{article}

    \usepackage[preprint,nonatbib]{neurips_2020}

\usepackage[utf8]{inputenc} 
\usepackage[T1]{fontenc}    
\usepackage{hyperref}       
\usepackage{url}            
\usepackage{booktabs}       
\usepackage{amsfonts}       
\usepackage{nicefrac}       
\usepackage{microtype}      

\usepackage{graphicx}
\usepackage{subfig}
\usepackage{amsmath}
\usepackage{amsthm}
\usepackage{subfiles}

\newtheorem{theorem}{Theorem}
\newtheorem{lemma}{Lemma}

\newtheorem{definition}{Definition}
\newtheorem*{lemma*}{Lemma}

\title{Towards Expressive Graph Representation}

\author{%
  Chengsheng Mao, Liang Yao, Yuan Luo \\
  Department of Preventive Medicine, Feinberg School of Medicine,\\
  Northwestern University\\
  \texttt{\{chengsheng.mao, liang.yao, yuan.luo\}northwestern.edu} \\
}

\begin{document}

\maketitle

\begin{abstract}
  Graph Neural Network (GNN) aggregates the neighborhood of each node into the node embedding and shows its powerful capability for graph representation learning. However, most existing GNN variants aggregate the neighborhood information in a fixed non-injective fashion, which may map different graphs or nodes to the same embedding, reducing the model expressiveness. We present a theoretical framework to design a continuous injective set function for neighborhood aggregation in GNN. Using the framework, we propose expressive GNN that aggregates the neighborhood of each node with a continuous injective set function, so that a GNN layer maps similar nodes with similar neighborhoods to similar embeddings, different nodes to different embeddings and the equivalent nodes or isomorphic graphs to the same embeddings. Moreover, the proposed expressive GNN can naturally learn expressive representations for graphs with continuous node attributes. We validate the proposed expressive GNN (ExpGNN) for graph classification on multiple benchmark datasets including simple graphs and attributed graphs. The experimental results demonstrate that our model achieves state-of-the-art performances on most of the benchmarks.
\end{abstract}

\section{Introduction}

Recently, graph data are utilized in more and more application and research domains. Graph neural networks (GNN) that can learn a distributed representation for a graph or a node in a graph are widely applied to a variety of areas for graph data analysis, such as social network analysis \cite{hamilton2017inductive,ying2018graph}, molecular structure inference \cite{duvenaud2015convolutional,gilmer2017neural}, text mining \cite{yao2019graph,peng2018large}, clinical decision making \cite{li2018classifying,mao2019medgcn} and image processing \cite{mao2019imagegcn,garcia2017few}, etc. GNN recursively updates the representation of a node in a graph by aggregating the feature vectors of its neighbors and itself \cite{hamilton2017inductive,morris2019weisfeiler,xu2019powerful}. The graph-level representation can then be obtained through aggregating the final representations of all the nodes in the graph. The generated representations can feed into a prediction model for different learning tasks, such as node classification and graph classification, and the whole model can be trained in an end-to-end fashion. 

\begin{figure*}
  \centering
  \includegraphics[width=.8\linewidth]{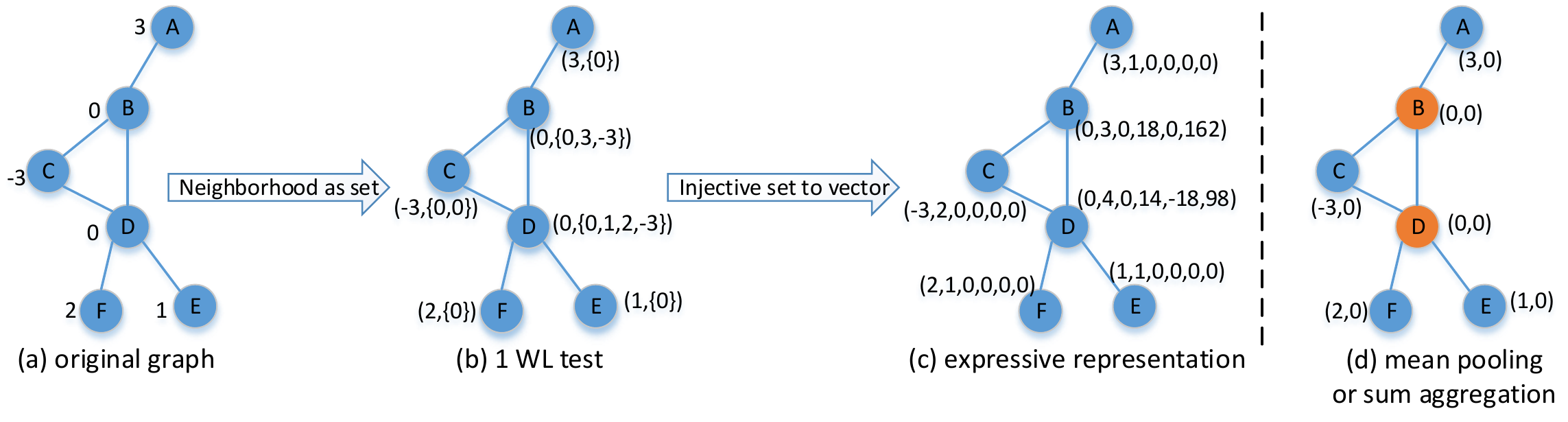}
  \caption{An overview of our framework on an exemplar attributed graph in one iteration. (a) Original graph with attributed nodes; (b) Graph nodes are represented by the corresponding attribute and neighborhood set through WL test; (c) The node vector representations after an injective set function on neighborhood sets, here the set function is $f(X)=\sum_{x \in X}(1,x, x^2, x^3, x^4)$; (d) A non-injective alternative of our injective set function in other GNNs, the node representations after mean pooling or sum aggregation of neighborhood sets. After aggregation, the node information remain unchanged, node B and D still have the same representation despite their different neighborhoods.}
  \label{fig:illustration}
\end{figure*}

In GNN, the aggregation rule plays a vital role in learning informative representations for the nodes and the entire graph. There are many GNN variants with different aggregation rules proposed to achieve good performances in different tasks, e.g., graph convolutional networks (GCN) \cite{kipf2017semi} and GraphSAGE \cite{hamilton2017inductive} for node-level aggregation and deep graph convolutional neural network (DGCNN) \cite{zhang2018end} and capsule neural network (CapsNet) \cite{xinyi2019capsule} for graph-level aggregation. However, most of the existing GNN aggregation rules are designed based on a fixed non-injective pooling function, e.g., max pooling and mean pooling. The fixed non-injective pooling usually loses some information and may generate the same embedding for different nodes or graphs. For example, for the graph with attributed nodes in Figure \ref{fig:illustration}(a), mean pooling or sum aggregation on the neighborhoods generates the same neighborhood representation for all the nodes (Figure \ref{fig:illustration}(d)), thus cannot capture any meaningful structure information. Xu et al. \cite{xu2019powerful} showed that a powerful GNN can at most achieve the discriminative power as Weisfeiler-Lehman graph isomorphism test (WL test) which can discriminate a broad class graphs \cite{weisfeiler1968reduction}, and proposed the powerful graph isomorphism network (GIN). However, the theoretical framework of GIN is under the assumption that the input feature space is countable, which makes GIN less expressive when applied to graphs with continuous attributes, i.e., attributed graphs.

Here, we present a theoretical framework that can guide us to design highly expressive GNNs for general graphs with continuous attributes. Our framework is illustrated in Figure \ref{fig:illustration}, first, each node with attribute in a graph (Figure \ref{fig:illustration}(a)) is converted to a tuple representation containing its attribute and a set of its neighborhood attributes through one WL test iteration (Figure \ref{fig:illustration}(b)), then we design a continuous injective set function to map each neighborhood to a vector (Figure \ref{fig:illustration}(c)). After certain learnable transformations, the graph with node embeddings can go the next iteration. After \textit{k} iterations, a node representation can capture the structural information and attribute information within the node’s \textit{k}-hop neighborhood. Due to the continuous injective set mapping, equivalent nodes in the graph have the same representation and vice versa. Our code is available at \url{ https://github.com/mocherson/Exp_GNN}.

Our main contributions are summarized as follows. (1) We present a theoretical framework to guide the design of highly expressive GNNs for general attributed graphs. (2) Using the framework, we develop two variants of ExpGNN with a customized continuous injective set function and a learned continuous set function, respectively. (3) We validate our models on multiple benchmark datasets including simple graphs and attributed graphs for graph classification, the experimental results demonstrate that our models can achieve state-of-the-art performances on most of the benchmarks. 

\section{Preliminaries}


A graph $G$ is denoted as $(V,E)$, where $V$ is the node set (assume size $n$) corresponding to a node feature matrix $X\in \mathbb{R}^{n\times d}$, and $E$ is the set of edges corresponding an adjacency matrix $A\in\{0,1\}^{n\times n}$. We focus on node features in this paper, and leave edge features for future work. 


\textbf{Graph neural networks.}
GNNs update a node representation by aggregating the node's neighborhood and then combine the neighborhood representation and the node's current representation. Formally, the propagation rule of a GNN layer can be represented as
\begin{equation}\label{eq:combine}
H^{(k+1)}(v) =    
f_{C}^{(k)} \left(H^{(k)}(v) , f_{A}^{(k)}\left( \left\{H^{(k)}(w) | w\in \mathcal{N}(v) \right\}\right) \right)  
\end{equation}
where $H^{(k)}(v)$ is the representation vector of node $v$ in the $k$th layer, and $H^{(0)}(v)$ is initialized with $X(v)$, the original attributes of node $v$. $\mathcal{N}(v)$ is the neighborhood of $v$. $f_{A}^{(k)}(\cdot)$ AGGREGATEs over a neighborhood and $f_{C}^{(k)}(\cdot)$ COMBINEs the node's current representation and its neighborhood’s representation in the $k$th layer.  

For node embedding, the node representation of the final layer $H^{(K)}(v)$ (suppose a total of $K$ layers) is considered as an informative representation that could be used for node classification. For graph or subgraph embedding, another READOUT function $f_{R}(\cdot)$ is employed to obtain the graph-level representation $h_G$ by aggregating the final representations of nodes in the graph or subgraph $G$, i.e.,

\begin{equation}\label{eq:readout}
H_G = f_{R} \left( \left\{H^{(K)}(v) | v\in G \right\}) \right)
\end{equation}

$f_{A}(\cdot)$, $f_{C}(\cdot)$ and $f_{R}(\cdot)$ are all crucial for the expressive capability of a GNN. $f_{A}(\cdot)$ and $f_{R}(\cdot)$ are set functions that maps a set to a vector, they can be simple summations or sophisticated graph-level pooling functions \cite{ying2018hierarchical,zhang2018end}. $f_{C}(\cdot)$ operates on two vectors, it can be usually modeled by a multi-layer perceptron (MLP) or linear function on the concatenated vector.

\textbf{The expressive capability of GNN.}
Lemma 2 and Theorem 3 in Xu et al.'s work \cite{xu2019powerful} describe the relation between GNNs and WL test in expressive capability for graphs. We outline them as follows, refer to \cite{xu2019powerful} for the proofs.   
\begin{lemma}
If the WL test decides two graphs $G_1$ and $G_2$ are isomorphic, any GNNs defined by Eq. \ref{eq:combine} and \ref{eq:readout} will map $G_1$ and $G_2$ to the same embedding.
\end{lemma}
\begin{theorem} \label{th:gnnpower}
If WL test decides two graphs $G_1$ and $G_2$ are not isomorphic, a GNN with sufficiently many GNN layers defined by Eq. \ref{eq:combine} and \ref{eq:readout} can also map $G_1$ and $G_2$ to different embeddings if the functions $f_{A}(\cdot)$, $f_{C}(\cdot)$ and $f_{R}(\cdot)$ are all injective.
\end{theorem}

The above Lemma and Theorem can guide us to design a GNN that has the discriminative power equal to WL test. The key is to design injective functions for $f_{A}(\cdot)$, $f_{C}(\cdot)$ and $f_{R}(\cdot)$. An injective function for $f_{C}(\cdot)$ that operates on two vectors can be easily obtained by concatenating the two vectors. But designing an injective function for $f_{A}(\cdot)$ or $f_{R}(\cdot)$ that operates on a set is not trivial, because a set can have variable number of elements, and the operation on the set elements must be permutation-invariant. Moreover, the continuity of these functions are all crucial to the model expressive capability, which is not considered in \cite{xu2019powerful}. In the following, we will discuss how to design the continuous injective aggregation function on a set and further expressive GNNs.

\section{Methods}

\subsection{Set representation}
A set function is a function defined in a domain that is a collection of sets. In a finite graph, the neighborhood of each node is considered as a finite set. Thus, in this paper, we only consider set functions of finite sets. A continuous set function is of real importance in practice \cite{wagstaff2019limitations}. Generally, in a continuous function, sufficiently small changes in the input result in arbitrarily small changes in the output at every point in the domain. The continuity of a function ensures that the change in output is very slight if the input is altered slightly by any reason such as truncating to machine precision. In this paper, we talk about the ordinary continuity where the continuity of function $f(\mathbf{x})$ at point $\mathbf{c}$ is defined by the limitation as $\lim_{\mathbf{x}\rightarrow \mathbf{c}} f(\mathbf{x}) = f(\mathbf{c})$

For $M\in \mathbb{N}$, a set function $f(X)$ defined in domain $\mathcal{X}=\{X | X\subset \mathbb{R}^d, |X|\leq M\}$ can be represented as a sequence of permutation-invariant functions $f_i$ for different set sizes, i.e., 
\begin{equation}\label{eq:setfunction}
f(X)=f_i(x_1,\cdots,x_i) \quad if \quad |X|=i\leq M, x_1,\cdots,x_i \in X
\end{equation}

\begin{definition}[Continuous set function]
For $M \in \mathbb{N}$ and a set function $f(X)$ defined in domain $\mathcal{X}_M=\{X \mid X \subset \mathbb{R}^d,|X| \le M\}$, $f(X)$ can be represented as Eq. \ref{eq:setfunction}, if $f_i(x_1,\cdots,x_i)$ is continuous in the Euclidean space for every $i\leq M$, we call $f(X)$ a continuous set function.
\end{definition}

Obviously, a continuous set function can also have the property that sufficiently small changes in the input (a sufficiently small change will not change the set size) result in arbitrarily small changes in the output. Thus by a continuous set function, graphs with very similar structures and attributes could be mapped to similar embeddings.

The following theorem provides a way to construct continuous injective set functions in uncountable space by sum aggregation after a certain transformation. 
\begin{theorem} \label{th:injective}
Assume $\mathcal{X}$ is a set of finite subsets of $\mathbb{R}^d$, i.e., for $M\in \mathbb{N}$ and  $\mathcal{X}=\{X|X\subset \mathbb{R}^d, |X|\leq M \}$, there exists an infinite number of continuous functions $\Phi : \mathbb{R}^d\rightarrow \mathbb{R}^{D}$ such that the set function $f:\mathcal{X}\rightarrow \mathbb{R}^{D}$, $f(X)=\sum_{\mathbf{x}\in X} \Phi(\mathbf{x})$ is continuous and injective.
\end{theorem}
We prove Theorem \ref{th:injective} in the supplementary material, the proof contains three steps: 1. Constructing a satisfying function in one dimensional cases ($d=1$); 2. Constructing a satisfying function in multi-dimensional cases($d>1$) based on the results in step 1; 3. The satisfying function can be used to generate infinite many other satisfying functions. 

In our proof, we find a $\Phi(\mathbf{x})$ defined in Eq. \ref{eq:phidd} ($\mathbf{x}[i]$ is the $i$th component of vector $\mathbf{x}$) that can make $f(X)=\sum_{\mathbf{x}\in X} \Phi(\mathbf{x})$ continuous and injective if the first entries of all vectors in $X$ are distinct.
\begin{equation}\label{eq:phidd}
\Phi_{M}(\mathbf{x})=  \\
\left[ 
\begin{matrix} 
    1, & \mathbf{x}[1], & \mathbf{x}[1]^2,  & \cdots, & \mathbf{x}[1]^{(M-1)}, & \mathbf{x}[1]^{M},   \\
    \mathbf{x}[2], & \mathbf{x}[1]\mathbf{x}[2], & \mathbf{x}[1]^2\mathbf{x}[2], & \cdots, & \mathbf{x}[1]^{M-1}\mathbf{x}[2],  \\
    \vdots  & \vdots & \vdots  & \ddots & \vdots \\
    \mathbf{x}[d], & \mathbf{x}[1]\mathbf{x}[d], & \mathbf{x}[1]^2\mathbf{x}[d], & \cdots, & \mathbf{x}[1]^{M-1}\mathbf{x}[d] 
    \end{matrix}
    \right] 
\end{equation}

In the proof, we also provide a way to construct such a function $\Phi(\mathbf{x})$ by defining a continuous injective function $g:\mathbb{R}^d\rightarrow \mathbb{R}^d$, then $\Phi(g(\mathbf{x}))$ can also satisfy the condition if we have a function $\Phi(\mathbf{x})$ satisfying the condition. We call $\Phi(\mathbf{x})$ the \textbf{transformation function}.

\begin{lemma} \label{lm:limitation}
Let $M\in \mathbb{N}$ and $\mathcal{X}=\{X | X \subset \mathbb{R}^d, |X|=M \}$, then for any continuous function $\Phi : \mathbb{R^d}\rightarrow \mathbb{R}^{N}$, if $N < dM$, the set function $f:\mathcal{X}\rightarrow \mathbb{R}^{N}$ $f(X)=\sum_{\mathbf{x}\in X} \Phi(\mathbf{x})$ is not injective.
\end{lemma}
We prove Lemma \ref{lm:limitation} in the appendix. Lemma \ref{lm:limitation} tells that if we want to construct a continuous injective set function for sets with $M$ $d$-dimensional vectors by sum aggregation with continuous transformation $\Phi(\mathbf{x})$, $\Phi(\mathbf{x})$ must have at least $dM$ dimensions. We are restricting $\Phi$ as a continuous function so that it can be modeled by a neural network, because a neural network can approximate any continuous functions rather than any functions by the universal approximation theorem \cite{cybenko1989approximation}.

\subsection{Expressive graph representation}
Since Theorem \ref{th:injective} tells that a set can be uniquely represented by a sum aggregation of its elements through a transformation function, we can use the unique set representation to model neighborhood of each node in a graph, and thus improve the expressiveness of graph representation. From Theorem \ref{th:gnnpower}, to design an expressive GNN, we need to design injective functions for $f_{A}(\cdot)$, $f_{C}(\cdot)$ and $f_{R}(\cdot)$. Since $f_{C}(\cdot)$ is easy to get continuous and injective, and $f_{A}(\cdot)$ and $f_{R}(\cdot)$ both operate on a set of vectors in $R^d$, thus, we need Theorem \ref{th:injective} to guide us to construct a continuous and injective set function for $f_{A}(\cdot)$ and $f_{R}(\cdot)$ by sum aggregation after a certain continuous transformation. 

\textbf{COMBINE function.}
According to Lemma \ref{lm:limitation}, for a set of $M$ $d$-dimensional embeddings, the transformation function must be at lease $dM$-dimensional to construct a continuous injective set function with sum aggregation, thus, without dimension reduction in $f_{C}(\cdot)$, we get a $(dM+d)$-dimensional embedding after one layer.
Nevertheless, in a specific learning task, not all dimensions are related to the learning task, we could design learnable neural networks to adaptively reduce the output dimension for each layer. We could use learnable MLPs with a lower output dimension to model $f_{C}(\cdot)$ as Eq. \ref{eq:MLPcombine}, where $[\mathbf{x_1},\mathbf{x_2}]$ is to concatenate vectors $\mathbf{x_1}$ and $\mathbf{x_2}$.
\begin{equation}\label{eq:MLPcombine}
 f^{(k)}_{C}(\mathbf{x_1},\mathbf{x_2}) = MLP^{(k)}\left(\left[\mathbf{x_1},\mathbf{x_2}\right]\right)
\end{equation}
Note that MLP maping high-dimensional vectors to low-dimensional vectors must not be continuous and injective if all dimensions in the high-dimensional vectors are independent. Here, MLP is used for task-driven feature reduction.

\textbf{AGGREGATE function.}
 $f_{A}(\cdot)$ operates on a set of node embeddings in the neighborhood of a node. We have two choice of the transformation function of $f_{A}(\cdot)$, i.e., \textit{fixed transformation} and \textit{learnable transformation}.

\textit{Fixed transformation.}
In the proof of Theorem \ref{th:injective}, we find the function $\Phi_{M}(\mathbf{x})$ defined in Eq. \ref{eq:phidd} can be used as a continuous transformation function to make the sum aggregation continuous and injective in most cases. For $f_{A}(\cdot)$, let $M_n$ be the max neighborhood size for all nodes in all the graphs. Usually, $M_n$ is not very large, we can set the transformation function as $\Phi_{M_n}(\mathbf{x})$ for each layer $k$ to maintain the expressive capability. Then $f_{A}(\cdot)$ for layer $k$ can be represented as
\begin{equation}\label{eq:poweragg}
f_{A}^{(k)} \left( \left\{H^{(k)}(w) | w\in N(v) \right\} \right) = \sum_{w\in \mathcal{N}(v)}{\Phi_{M_n} \left(H^{(k)}(w) \right)}
\end{equation}

We are aware of that $f_{A}^{(k)}(\cdot)$ in Eq. \ref{eq:poweragg} are not totally injective if the first entry of elements of $\mathcal{N}(v)$ are not all distinct. We think $f_{A}^{(k)}(\cdot)$ can capture much more information than many other known aggregation methods. Combining Eqs. \ref{eq:combine}, \ref{eq:MLPcombine} and \ref{eq:poweragg}, we get the propagation rule as 
\begin{equation}\label{eq:powerGNN}
H^{(k+1)}(v) = MLP^{(k)} \left( \left[H^{(k)}(v), \sum_{w\in \mathcal{N}(v)}{\Phi_{M_n} \left(H^{(k)}(w)\right)}\right] \right)
\end{equation}

Though the function $\Phi_{M}(\mathbf{x})$ defined in Eq. \ref{eq:phidd} can make the sum aggregation continuous and injective, it may result in numerical stability since the item $\mathbf{x}[1]^M$ will make the number become very large or very close to 0 if $M$ is very large. To address this issue, we use a continuous and injective function $g(\mathbf{x})$ to normalize the power, since in the proof of Theorem \ref{th:injective} we know $\Phi_{M}(g(\mathbf{x}))$ is also a qualified transformation function to make the sum aggregation continuous and injective, if $g(\mathbf{x})$ is continuous and injective. In this paper, we set 
\begin{equation}\label{eq:gx}
\begin{aligned}
& g(\mathbf{x})[1]=
\begin{cases}
\mathbf{x}[1]^{1/M},  & \mathbf{x}[1]\ge 0  \\
-(-\mathbf{x}[1])^{1/M}, & \mathbf{x}[1]<0 \\
\end{cases}  \\
& g(\mathbf{x})[2:d]= \mathbf{x}[2:d]
\end{aligned}
\end{equation}

\textit{Learnable transformation.} 
Due to the continuity of the transformation function, we can also set a learnable MLP to approach the transformation function for $f_{A}(\cdot)$ by the universal approximation theorem \cite{cybenko1989approximation}, then we get the propagation rule as
\begin{equation}\label{eq:MLPagg}
H^{(k+1)}(v)=   
MLP_{c}^{(k)} \left( \left[H^{(k)}(v), \sum_{w\in \mathcal{N}(v)}{MLP_{t}^{(k)}\left(H^{(k)}(w)\right)}\right] \right)
\end{equation}
where $MLP_{t}^{(k)}$ and $MLP_{c}^{(k)}$ serve as the transformation function and the combine function for the $k$th layer, respectively.

By this way, we get the all the node embeddings for all graphs. For node classification task, the final layer output node embedding can be input to an learnable MLP classifier to get the probability for each class. For graph or subgraph classification, we need another aggregation function $f_{R}(\cdot)$ to aggregate all the node embeddings in a graph.

\textbf{READOUT function.}
$f_{R}(\cdot)$ operates on a set of all node embeddings in a graph. Let $M_G$ be the max node number of all the graphs, for $d$-dimensional node embeddings, the continuous transformation function must be at least $dM$ dimensions for injective continuous $f_{R}(\cdot)$, thus generate a $(dm)$-dimensional graph-level embedding. For large graphs with many nodes, the output dimension will be very high. To avoid high-dimensional embeddings, we also use a learnable MLP as the transformation function to reduce the output dimension. 
\begin{equation}\label{eq:readoutdecom}
H_G = f_{R} \left( \left\{H^{(K)}(v) | v\in G \right\} \right) = \sum_{v\in G}{MLP_G \left(H^{(K)}(v)\right)}
\end{equation}
For graph classification, the output graph-level embedding are input to an MLP classifier with $n_C$ outputs corresponding to the probabilities of the $n_C$ classes. Also, MLP can represent the composition of functions, in our implementation, we merge the classifier MLP and MLP$_G$ into only one MLP as GIN did in \cite{xu2019powerful}. We only use the final GNN layer outputs for classification rather than concatenating all layers' outputs to construct a longer vector representation for classification as GIN did, because we think the final layer outputs contain all information from middle layers and are expressive enough for graph classification. In addition, this can reduce the input dimension of the final classifier, resulting in a simpler classifier than GIN, especially in case of many layers.  

\section{Related Work}
\textbf{General graph neural networks.}
Many GNN variants with different aggregation rules are proposed in the literature to achieve good performances in different tasks. In \cite{xu2019powerful}, GIN has been proposed with the propagation rule as 
\begin{equation}\label{eq:gin}
   H^{(k+1)}(v) =  
   MLP^{(k)}\left(\left(1+\epsilon^{(k)} \right)\cdot H^{(k)}(v) + \sum_{u\in N(u)}{H^{(k)}(u)} \right)
\end{equation}
GIN is expected to be highly expressive for simple graphs where node attributes can be one-hot encoders on which sum aggregation is injective. However, GIN cannot be directly extended to attributed graphs with the same expressive capability, because the sum aggregation is no longer injective in uncountable cases.

GCN is another GNN variant with simple element-wise mean pooling in a node's neighborhood including the node itself \cite{kipf2017semi}. Hamilton et al. \cite{hamilton2017inductive} tested 3 aggregators in GraphSAGE, including mean aggregator, LSTM aggregator and max pooling aggregator, they found no significant performance difference exists between the LSTM aggregators and pool aggregators, but GraphSAGE-LSTM is significantly slower than GraphSAGE-pool. Mean aggregation and max pooling are permutation invariant on sets, but the operation is not injective, which may result in the same embedding for different inputs. LSTM aggregation could have large expressive capacity, but it is not permutation invariant, this may render equivalent nodes or isomorphic graphs to have different embeddings.    

\textbf{Graph kernels for graph classification.} 
Graph kernels is an established and widely-used technique for solving classification tasks on graphs \cite{shervashidze2011weisfeiler,shervashidze2009efficient,yanardag2015deep}. One of the dominating paradigms in the design of graph kernels is representation and comparison of local structure by neighborhood aggregation. The well know WL subtree kernel inspired the GNN for neighborhood aggregation. Though most of the graph kernels are for simple graph classification, more and more research began to study graph kernels for attributed graphs \cite{morris2016faster,feragen2013scalable,orsini2015graph,neumann2016propagation}. Graph kernel-based method can measure the similarity between two graphs, but usually cannot generate a distributed representation for a graph.

\section{Experiments}
For ExpGNN, we evaluate the expressive capability on the training data and evaluate the generalization ability on the test data. The evaluations are based on two graph classification tasks, simple graph classification and attributed graph classification.

\begin{table*}[t]
\scriptsize
  \centering
  \caption{Accuracy for simple graph classification in test set (\%). Top 3 performances on each dataset are bolded.}
    \begin{tabular}{l|c|c|c|c|c|c|c|c}
    \toprule
          & MUTAG & PTC   & NCI1  & PROTEINS & COLLAB & IMDB-B & IMDB-M & RDT-B \\
    \midrule
    ExpGNN-fixed & \textbf{90.5 $\pm$ 6.1} & 65.6  $\pm$ 7.8 &  \textbf{82.9 $\pm$ 2.5}  &  \textbf{77.2  $\pm$5.6}  &  --     &   --    &   --    & -- \\
    ExpGNN-MLP & \textbf{91.1 $\pm$ 7.9}  &  \textbf{66.5 $\pm$  7.6}  &  \textbf{82.9 $\pm$ 1.3}  &  76.1 $\pm$ 5.2  &   --    &    --   &    --   & -- \\
    ExpGNN-FI-fixed &  90.0 $\pm$ 6.9  &  \textbf{68.5  $\pm$ 8.5}  &  82.3 $\pm$ 2.6  &  \textbf{77.0 $\pm$ 6.0}  &  --     &  73.5 $\pm$ 4.3  &  48.9 $\pm$ 3.3  & -- \\
    ExpGNN-FI-MLP &  90.0 $\pm$  6.9 &  64.1$\pm$  4.1  &  \textbf{83.3 $\pm$ 1.7}  &  76.3$\pm$  5.1  &  78.4$\pm$  1.0  &  73.3 $\pm$ 3.7  &  49.9 $\pm$ 2.8  &  \textbf{91.2  $\pm$4.2}  \\
    \midrule
    GIN-final &  89.4 $\pm$ 5.8 &  63.5 $\pm$  8.6  &  82.7 $\pm$ 2.0  &  76.2 $\pm$ 4.9  &  75.8 $\pm$ 1.8  &  72.9 $\pm$ 5.3  &  48.9 $\pm$ 4.9  &  \textbf{91.6 $\pm$  3.0}  \\
    GIN \cite{xu2019powerful}  & 89.4 $\pm$  5.6 & 64.6  $\pm$ 7.0 & 82.7  $\pm$ 1.6 & 76.2  $\pm$ 2.8 & \textbf{80.2 $\pm$  1.9} & \textbf{75.1 $\pm$  5.1} & \textbf{52.3 $\pm$ 2.8} & \textbf{92.4 $\pm$  2.5} \\
    GCN \cite{kipf2017semi}   & 85.6 $\pm$ 5.8 & 64.2 $\pm$ 4.3 & 80.2 $\pm$ 2.0 & 76.0 $\pm$ 3.2 & 79.0 $\pm$ 1.8 & 74.0 $\pm$ 3.4 & \textbf{51.9 $\pm$ 3.8} & 50.0 $\pm$ 0.0 \\
    GraphSAGE \cite{hamilton2017inductive} & 85.1 $\pm$ 7.6 & 63.9 $\pm$ 7.7 & 77.7 $\pm$ 1.5 & 75.9 $\pm$ 3.2 &    --   & 72.3 $\pm$ 5.3 & 50.9 $\pm$ 2.2 & -- \\
    PSCN \cite{niepert2016learning}  & \textbf{92.6 $\pm$ 4.2} & 60.0$\pm$ 4.8 & 78.6 $\pm$ 1.9 & 75.9 $\pm$ 2.8 & 72.6 $\pm$ 2.2 & 71.0 $\pm$ 2.2 & 45.2$\pm$ 2.8 & 86.3 $\pm$ 1.6 \\
    DCNN \cite{atwood2016diffusion} & 67.0  & 56.6 & 62.6 & 61.3 & 52.1 & 49.1  & 33.5  & -- \\
    DGCNN \cite{zhang2018end} & 85.8 $\pm$ 1.7 & 58.6 $\pm$ 2.5 & 74.4 $\pm$ 0.5 & 75.5 $\pm$ 0.9 & 73.8 $\pm$ 0.5 & 70.0 $\pm$ 0.9 & 47.8 $\pm$ 0.9 & -- \\
    CapsGNN \cite{xinyi2019capsule} & 86.7 $\pm$ 6.9 & -     & 78.4 $\pm$ 1.6 & 76.3 $\pm$ 3.6 & \textbf{79.6 $\pm$ 0.9} & 73.1 $\pm$ 4.8 & 50.3 $\pm$ 2.7 & -- \\
    GCAPS-CNN \cite{verma2018graph} & --     & \textbf{66.0 $\pm$ 5.9} & 82.7 $\pm$ 2.4 & \textbf{76.4 $\pm$ 4.2} & 77.7 $\pm$ 2.5 & 71.7 $\pm$ 3.4 & 48.5 $\pm$ 4.1 & 87.6 $\pm$ 2.5 \\
    IEGN \cite{Maron2019InvariantAE} & 84.6 $\pm$ 10 & 59.5 $\pm$ 7.3 & 73.7 $\pm$ 2.6 & 75.2 $\pm$ 4.3 & 77.9 $\pm$ 1.7 & 71.3 $\pm$ 4.5 & 48.6 $\pm$ 3.9 & -- \\
    HO-GNN \cite{morris2019weisfeiler} & 86.1  & 60.9  & 76.2  & 75.9  &   --    & \textbf{74.2}  & 49.5  &  --\\
    FGSD \cite{verma2017hunt} & 92.1 & 62.8 & {79.8} & {73.4} & \textbf{80.0} & {73.6} & {52.4} & 86.5 \\
    AWE \cite{ivanov2018anonymous}  & 87.9 $\pm$ 9.8 & --     & --     & --     & 73.9 $\pm$ 1.9 & \textbf{74.5 $\pm$ 5.9} & \textbf{51.5} $\pm$ 3.6 & 87.9 $\pm$ 2.5 \\
    Graph2vec \cite{narayanan2017graph2vec} & 83.2 $\pm$ 9.3 & 60.2 $\pm$ 6.9 & 73.2 $\pm$ 1.8 & 73.3 $\pm$ 2.1 &  --     & --      &  --     &--  \\
    \midrule
    WL subtree \cite{shervashidze2011weisfeiler}   & 90.4 $\pm$ 5.7 & 59.9 $\pm$ 4.3 & \textbf{86.0 $\pm$ 1.8} & 75.0 $\pm$ 3.1 & 78.9 $\pm$ 1.9 & 73.8 $\pm$ 3.9 & 50.9 $\pm$ 3.8 & 81.0 $\pm$ 3.1 \\
    GK \cite{shervashidze2009efficient}   & 81.6 $\pm$ 2.1 & 57.3 $\pm$ 1.4 & 62.5 $\pm$ 0.3 & 71.7 $\pm$ 0.6 & 72.8 $\pm$ 0.3 & 65.9 $\pm$ 1.0 & 43.9 $\pm$ 0.4 & 77.3 $\pm$ 0.2 \\
    DGK \cite{yanardag2015deep}  & 87.4 $\pm$ 2.7 & 60.1 $\pm$ 2.6 & 80.3 $\pm$ 0.5 & 75.7 $\pm$ 0.5 & 73.1 $\pm$ 0.3 & 67.0 $\pm$ 0.6 & 44.6 $\pm$ 0.5 & 78.0 $\pm$ 0.4 \\
    \bottomrule
    \end{tabular}%
  \label{tab:simplegraph}%
\end{table*}%

\begin{figure}[t]
  \centering
  \subfloat[MUTAG]{\includegraphics[width=.25\linewidth,page=1]{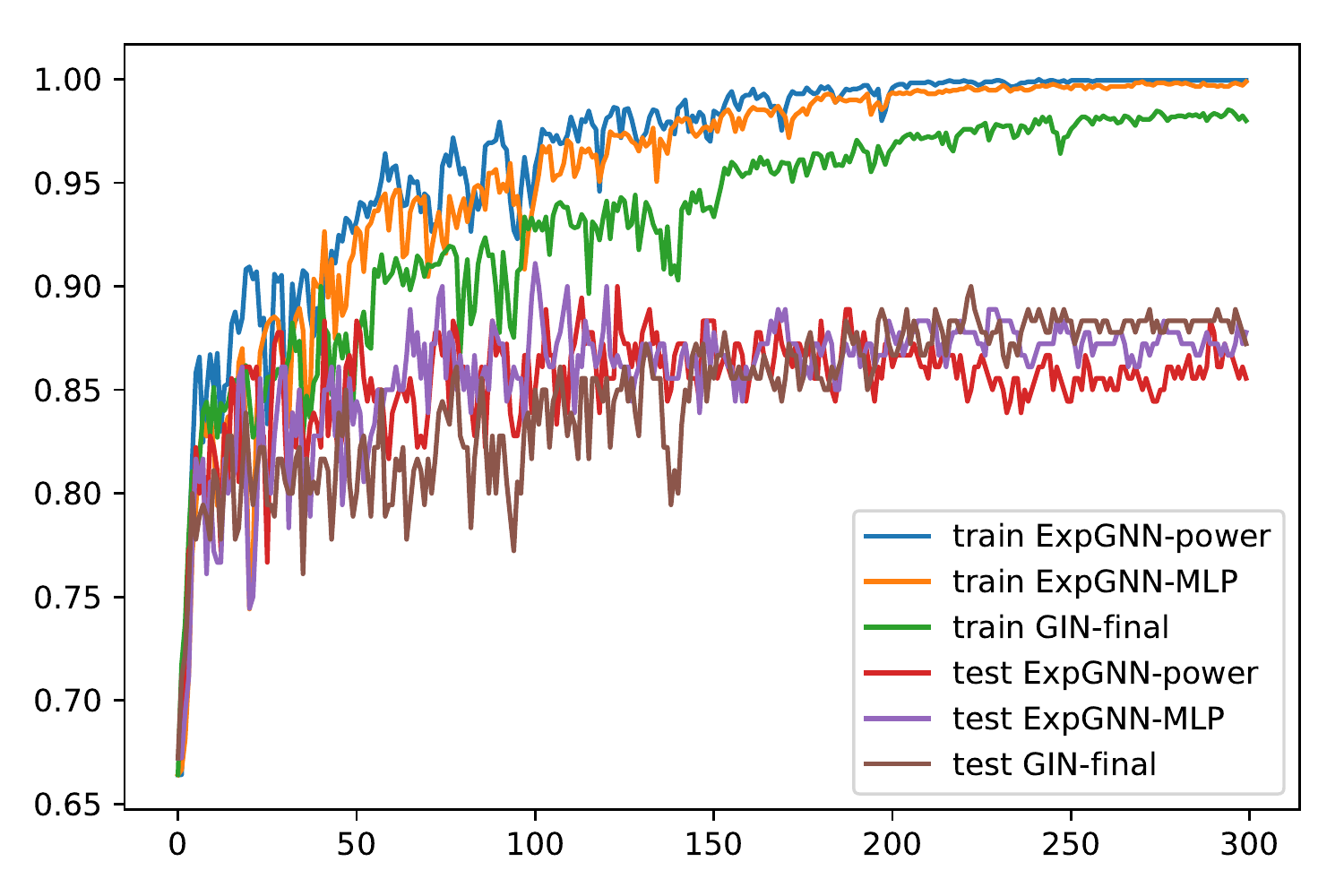}}
  \subfloat[PTC]{\includegraphics[width=.25\linewidth,page=1]{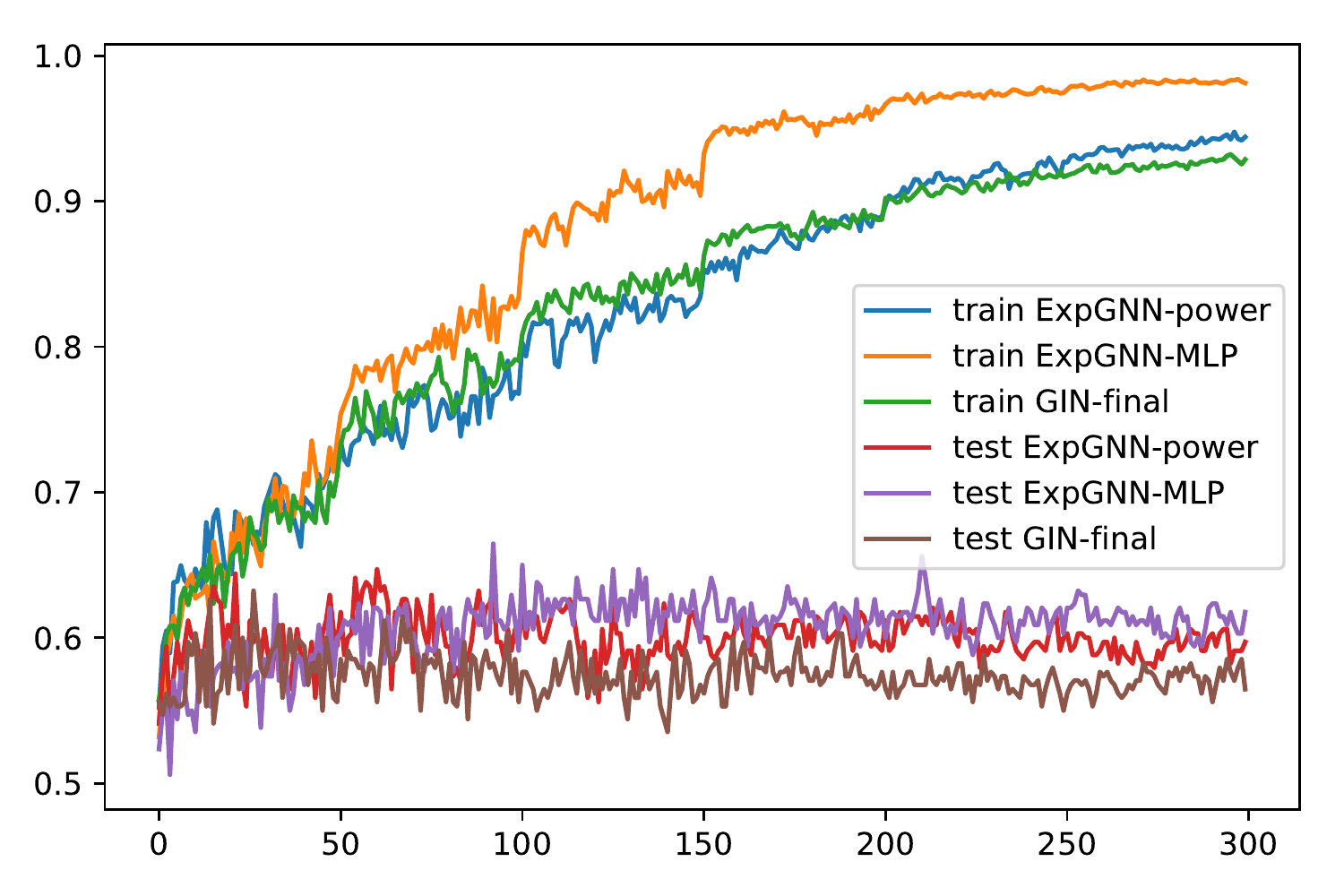}}
  \subfloat[PROTEINS-att]{\includegraphics[width=.25\linewidth,page=1]{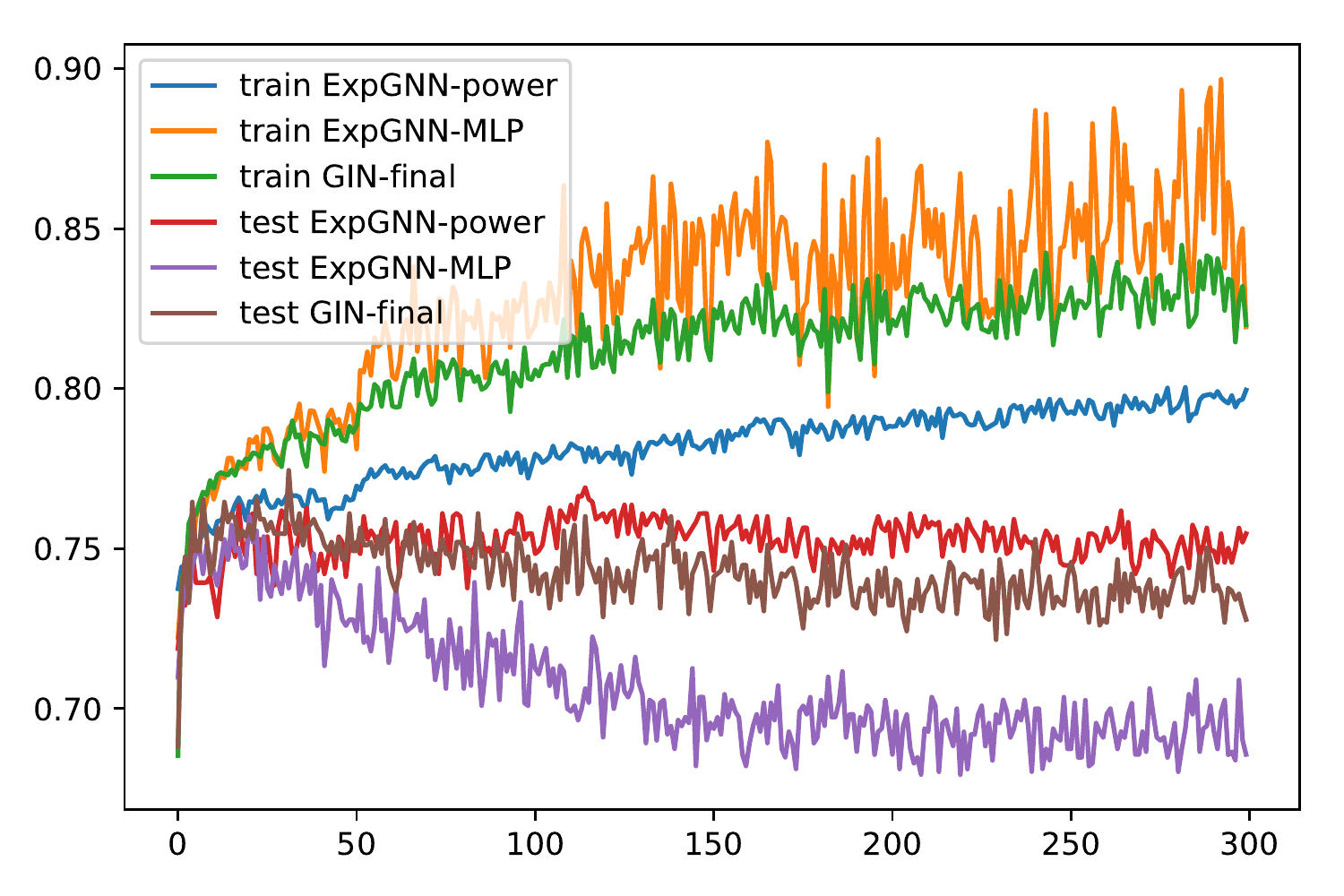}}
  \subfloat[FRANKENSTEIN]{\includegraphics[width=.25\linewidth,page=1]{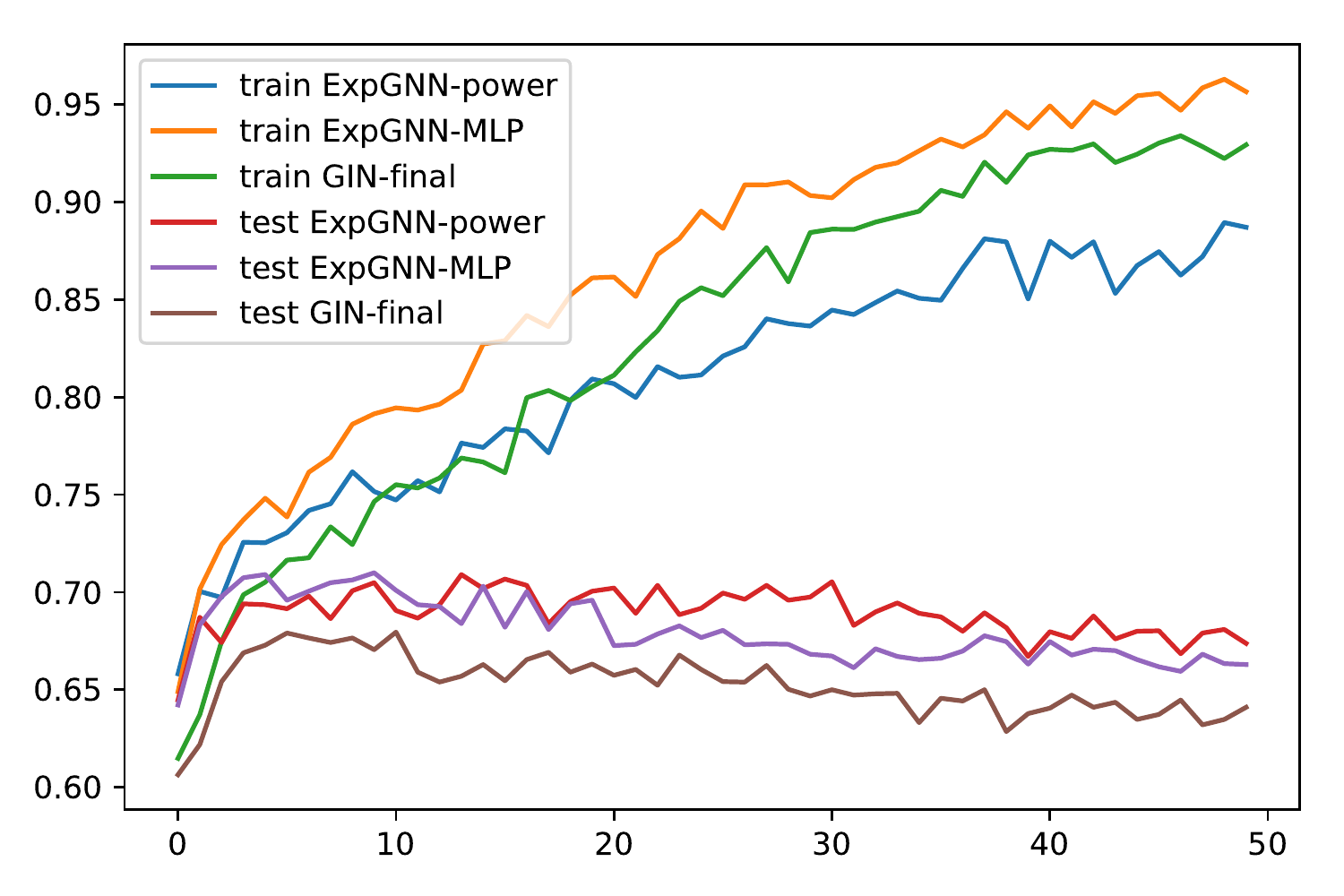}}
  
  
  
  \caption{The accuracy curves on training set and test set in the training process.}
  \label{fig:traincurve}
\end{figure}

\begin{table*}[t]
\scriptsize
  \centering
  \caption{Accuracy for attributed graph classification in test set (\%). Top 3 performances on each dataset are bolded.}
    \begin{tabular}{l|c|c|c|c|c}
    \toprule
          & ENZYMES & FRANKENSTEIN & PROTEINS-att & SYNTHETICnew & Synthie \\
    \midrule
    ExpGNN-fixed & 58.0 $\pm$ 5.6 & 71.5 $\pm$ 2.7 & \textbf{77.3 $\pm$ 3.2} & 89.7 $\pm$ 7.1 & 93.7 $\pm$ 3.7 \\
    ExpGNN-MLP & \textbf{71.3 $\pm$ 4.3} & \textbf{71.8  $\pm$ 2.7} & 76.1 $\pm$ 3.6 & \textbf{98.0 $\pm$ 2.7} & \textbf{99.8 $\pm$ 0.8} \\
    \midrule
    GIN-final & 69.3 $\pm$ 5.3 & 68.5 $\pm$ 1.5 & 76.4 $\pm$ 3.0  & 79.7 $\pm$ 6.9 & 90.0 $\pm$ 4.4\\
    HGK-SP \cite{morris2016faster} & \textbf{71.30 $\pm$ 0.86} & 70.06  $\pm$ 0.32 & \textbf{77.47  $\pm$ 0.43} & \textbf{96.46 $\pm$ 0.61} & 94.34 $\pm$0.54 \\
    HGK-WL \cite{morris2016faster} & 67.63 $\pm$ 0.95 & \textbf{73.62  $\pm$ 0.38} & 76.70 $\pm$ 0.41 & \textbf{98.84 $\pm$ 0.29} & \textbf{96.75 $\pm$ 0.51} \\
    GHK \cite{feragen2013scalable}  & 68.80 $\pm$ 0.96 & 68.48 $\pm$ 0.26 & 72.26 $\pm$ 0.34 & 85.10 $\pm$ 1.04 & 73.18 $\pm$ 0.77 \\
    GIK \cite{orsini2015graph}  & \textbf{71.70 $\pm$ 0.79} & \textbf{76.31 $\pm$ 0.33} & \textbf{76.88 $\pm$ 0.47} & 83.07  $\pm$ 1.10 & \textbf{95.75 $\pm$ 0.50} \\
    P2K \cite{neumann2016propagation}  & 69.22 $\pm$ 0.34 &   --    & 73.45 $\pm$ 0.48 & 91.70 $\pm$ 0.86 & 50.15 $\pm$ 1.92 \\
    \bottomrule
    \end{tabular}%
  \label{tab:resatt}%
\end{table*}%

\begin{figure*}[t]
  \centering
  \subfloat[ExpGNN-MLP]{\includegraphics[width=.25\textwidth,page=1]{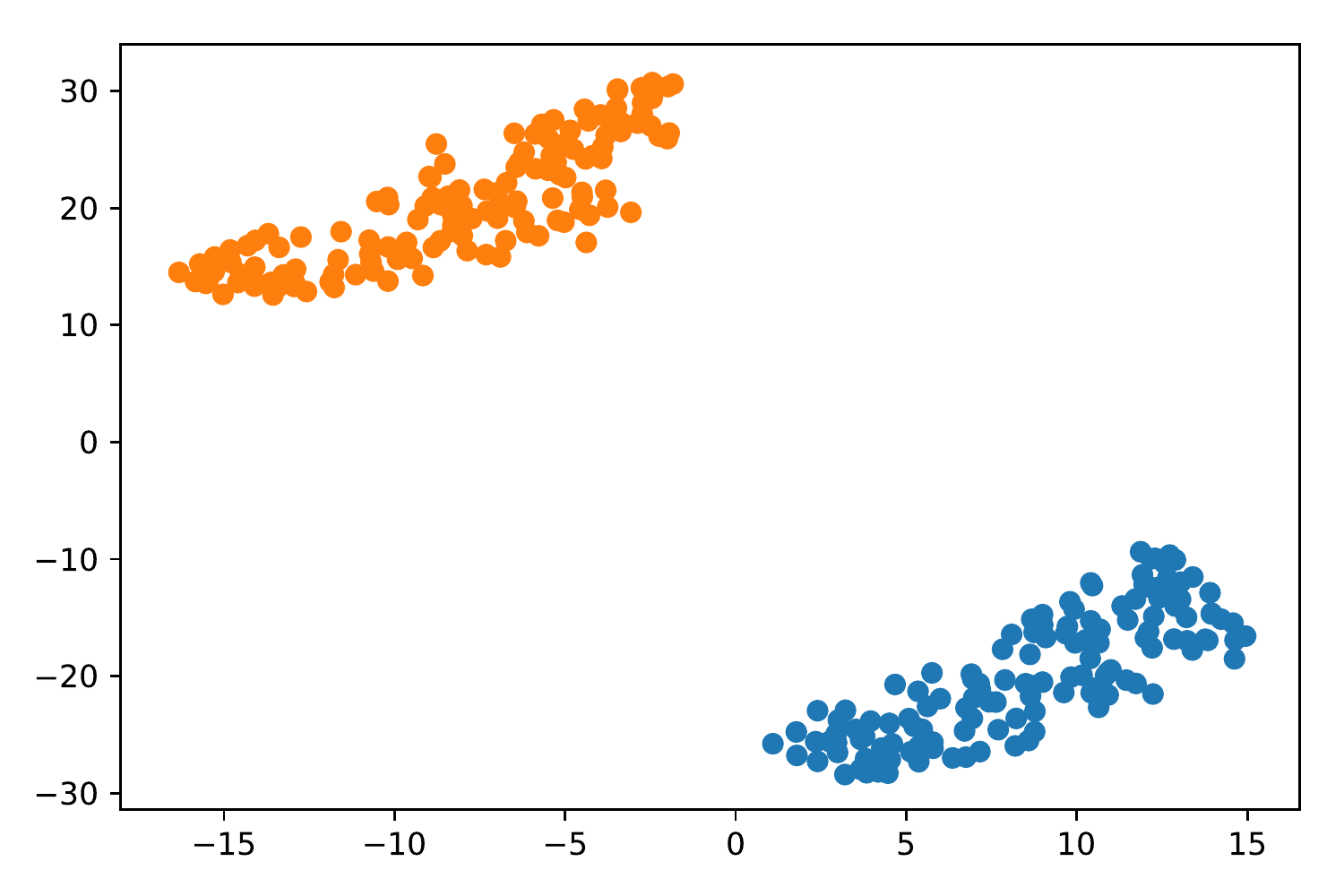}}
  \subfloat[ExpGNN-fixed]{\includegraphics[width=.25\textwidth,page=1]{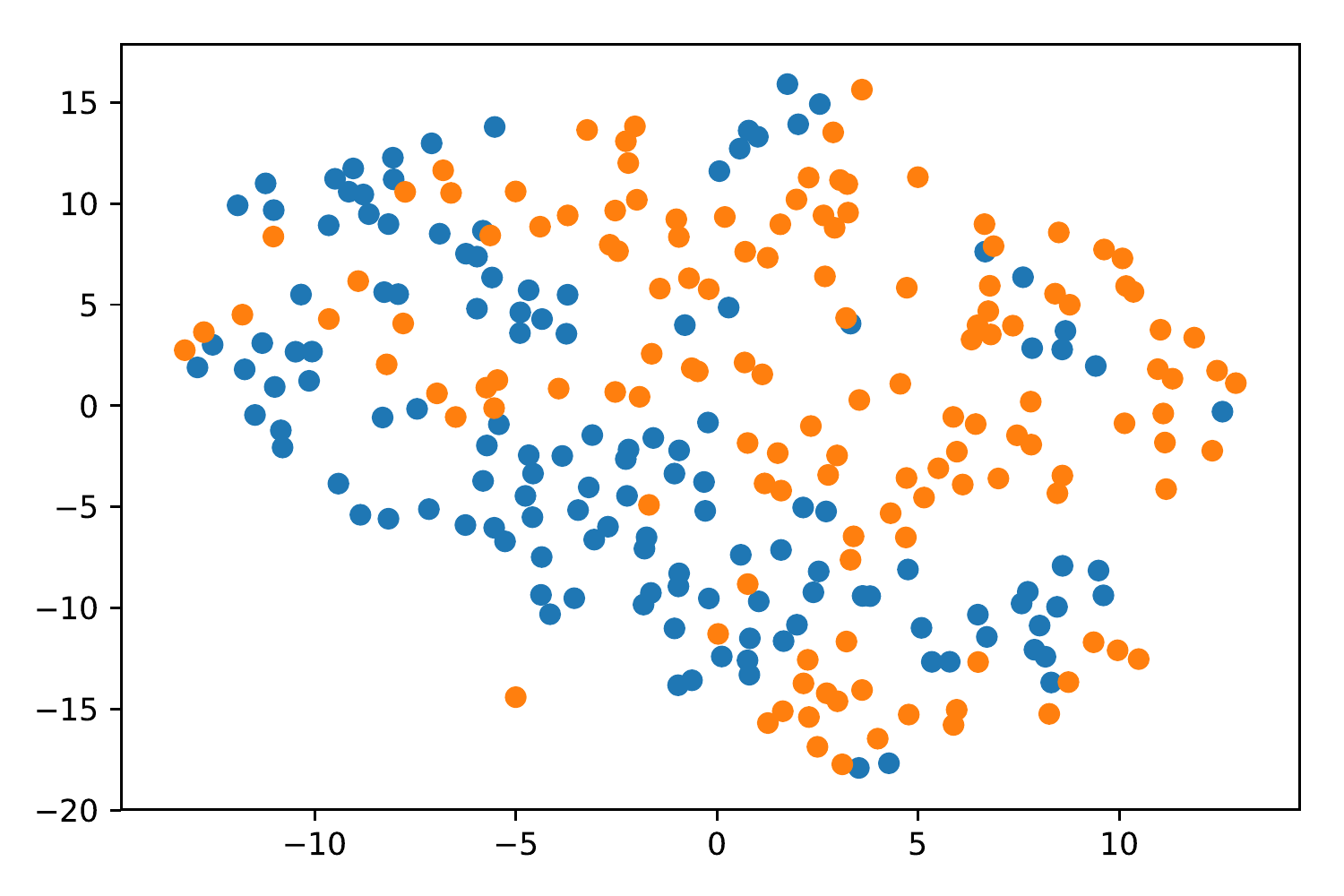}}
  \subfloat[GIN-final]{\includegraphics[width=.25\textwidth,page=1]{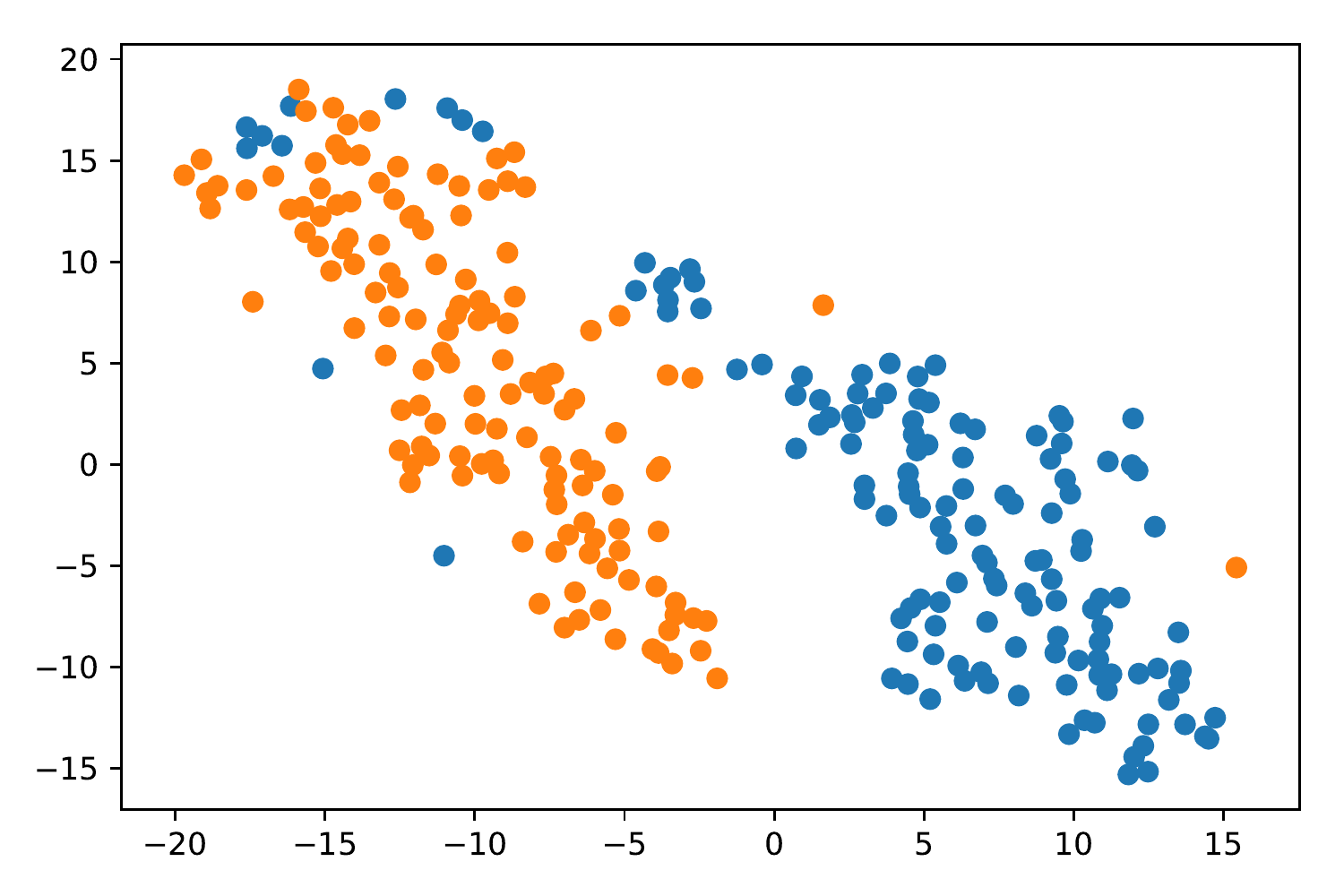}}
  \caption{t-SNE visualization of the output embeddings on training data of SYNTHETICnew dataset.}
  \label{fig:tsnevisual}
\end{figure*}



\subsection{Graph classification}
\textbf{Dataset.} We use 8 simple graph benchmarks and 5 attributed graph benchmarks for graph classification, the 8 simple graph datasets contain 4 bioinformatics datasets (MUTAG, PTC, NCI1, PROTEINS) and 4 social network datasets (COLLAB, IMDB-BINARY, IMDB-MULTI, and REDDIT-BINARY) \cite{yanardag2015deep}, these datasets have no continuous node attributes input to the model, for bioinformatics datasets, the categorical node labels are encoded as one-hot input features; for social network datasets, because nodes have no given features, we initialize all node features to 1. The 5 attributed graph datasets contain 3 bioinformatics datasets (ENZYMES, FRANKENSTEIN, PROTEINS-att) and 2 synthetic datasets (SYNTHETICNEW, Synthie) where the continuous attributes are concatenated with one-hot node label for each node to serve as the input to the models. All the datasets are available from \cite{KKMMN2016}. More dataset information can be found in the supplementary material. 

\textbf{Settings.} We implement 4 ExpGNN variants: (1) ExpGNN-fixed, the transformation function in all layers are set as Eq. \ref{eq:phidd}; (2) ExpGNN-MLP, the transformation function in all layers are set as a learnable MLP. (3) ExpGNN-FI-fixed and (4) ExpGNN-FI-MLP, because for a simple graph with one-hot node features, the summation without transformation or with identical transformation is injective, thus we set First layer Identical transformation function in ExpGNN-FI-fixed and ExpGNN-FI-MLP and only for simple graph classification. We also implement GIN with the output of the final layer as node embeddings to sum to graph embedding, GIN-final. For social network datasets, because the input nodes' features are all the same, any transformation function will generate the same node vector for all nodes, no transformation function is needed in the first layer, only FI is implemented. For COLLAB and REDDITBINARY, the max neighborhood size is too large, we do not implement fixed transformation function.

All the 4 ExpGNN variants and GIN-final have 5 GNN layers, all MLPs in ExpGNN have 2 layers. Batch normalization \cite{ioffe2015batch} is applied in every hidden layer (including GNN layer and MLP layer) followed by a ReLU activation function. We use the Adam optimizer \cite{kingma2014adam} with initial learning rate tuned in $\{0.01, 0.001\}$ and decay the learning rate by 0.5 every 50 epochs. The batch size is 32, no dropout layer applied. The number of hidden units is tuned in $\{16,32,64\}$. Following the settings in \cite{xu2019powerful,yanardag2015deep,niepert2016learning}, we perform 10-fold cross-validation on each dataset, and report the average and standard deviation of validation accuracies across the 10 folds within the cross-validation to evaluate the generalization ability. All models are trained 300 epochs, a single epoch with the best cross-validation accuracy averaged over the 10 folds is selected. To evaluation the expressive capability, we also record the average training accuracy across the 10 folds of ExpGNNs and GIN-final in each epoch.

\textbf{Baselines.}
For expressive capability, we compare ExpGNN with GIN-final on the training set. For generalization ability in simple graph classification, besides GIN, we also compare ExpGNN with a number of state-of-the-art models, including deep learning-based methods and graph kernels. The deep learning-based methods include GCN \cite{kipf2017semi}, GraphSAGE \cite{hamilton2017inductive}, PATCHY-SAN (PSCN) \cite{niepert2016learning}, Diffusion convolutional neural networks (DCNN) \cite{atwood2016diffusion}, Deep Graph CNN (DGCNN) \cite{zhang2018end}, Capsule Graph Neural Network (CapsGNN) \cite{xinyi2019capsule}, Graph capsule CNN (GCAPS-CNN) \cite{verma2018graph}, Invariant and equivariant graph networks (IEGN) \cite{Maron2019InvariantAE}, Higher-order Graph Neural Networks (HO-GNN) \cite{morris2019weisfeiler}, family of graph spectral distances (FGSD) \cite{verma2017hunt}, Anonymous Walk Embeddings (AWE) \cite{ivanov2018anonymous} and Graph2vec  \cite{narayanan2017graph2vec}. The graph kernel methods include WL subtree kernel \cite{shervashidze2011weisfeiler}, graphlet count kernel (GK) \cite{shervashidze2009efficient} and Deep Graph Kernel (DGK) \cite{yanardag2015deep}. Although deep learning methods can naturally handle attributed graphs, few results of attributed graph classification with deep learning methods are available in the literature. For generalization ability in attributed graph classification, we are only aware of graph kernel related baselines, including hash graph kernels with shortest path (HSK-SP) \cite{morris2016faster}, hash graph kernels with WL subtree (HSK-WL) \cite{morris2016faster}, GraphHopper kernel (GHK) \cite{feragen2013scalable}, graph invariant kernel (GIK) \cite{orsini2015graph} and propagation kernel for continuous attributes (P2K) \cite{neumann2016propagation}. We also compared ExpGNN with our implemented GIN-final for attributed graph classification.

\textbf{Results.}
Figure \ref{fig:traincurve} illustrates accuracies in training and test sets in the training process on 4 datasets for ExpGNN and GIN-final with their respective best settings. More results on other datasets can be found in the supplementary material. We can see that ExpGNN-MLP for different datasets are able to fit the training sets perfectly and are better than GIN-final and ExpGNN-fixed. For MUTAG and PTC datasets, GIN in \cite{xu2019powerful} can fit the training set, while GIN-final cannot, since GIN concatenates all the middle layer outputs as the graph embedding, it may be the reason that the final layer outputs of GIN may lose some information from middle layers. From the test accuracy curves in Figure \ref{fig:traincurve}, we can also find that ExpGNN can generalize better than GIN-final on these datasets, except ExpGNN-MLP on dataset PROTEINS-att. 

As for the generalization ability, Table \ref{tab:simplegraph} lists the classification accuracies for simple graph classification, Table \ref{tab:resatt} lists the results for attributed graph classification, comparing with other state-of-the-art methods. We highlight the top 3 accuracies for each dataset in boldface. From Table \ref{tab:simplegraph}, for simple graph classification on all bioinformatics datasets, at least 2 ExpGNN variants can achieve top 3 in these 21 models. From Table \ref{tab:resatt}, for all the attributed graph datasets, ExpGNN can achieve top 3 in these 8 models, especially, ExpGNN-MLP places first on Synthie dataset. Comparing ExpGNN and GIN-final, ExpGNN can consistently outperform GIN-final except for ExpGNN-fixed on ENZYMES dataset and ExpGNN-MLP on PROTEINS-att.

\subsection{Expressive capability analysis}
The expressive capability describes how a model can distinguish different samples. Generally, a high expressive model will map different samples to different embeddings and similar samples to similar embeddings. For classification problem, an expressive model should make samples in the same class compact together and samples in different classes highly discriminative. Here, we fetch the output embeddings of GNN before feeding to the classifier, and visualize them to see if GNN can discriminate samples from different classes. Figure \ref{fig:tsnevisual} shows the t-SNE visualization \cite{maaten2008visualizing} of output embeddings of different GNN models on training data of SYNTHETICnew dataset. The final layer GNN output embedding is also visualized with PCA and two random dims in the supplementary material. We can see that the output embeddings of ExpGNN-MLP are highly discriminative, which demonstrates the expressive capability of ExpGNN-MLP.

\section{Conclusion}
In this paper, we present a theoretical framework to design highly expressive GNNs for general graphs. Based on the framework, we propose two expressive GNN variants with fixed transformation function and learnable transformation function, respectively. Moreover, the proposed expressive GNN can naturally learn expressive representations for graphs with continuous node attributes. We validate the proposed GNN for graph classification on multiple benchmark datasets including simple graph and attributed graph. The experimental results demonstrate that our model achieves state-of-the-art performances on most of the benchmarks. 


\bibliographystyle{plain}
\bibliography{references}


\end{document}


\maketitle

\section{Proofs}
\begin{theorem} \label{th:injective}
Assume $\mathcal{X}$ is a set of finite subsets of $\mathbb{R}^d$, i.e., for $M\in \mathbb{N}$ and  $\mathcal{X}=\{X|X\subset \mathbb{R}^d, |X|\leq M \}$, there exists an infinite number of continuous functions $\Phi : \mathbb{R}^d\rightarrow \mathbb{R}^{D}$ such that the set function $f:\mathcal{X}\rightarrow \mathbb{R}^{D}$, $f(X)=\sum_{\mathbf{x}\in X} \Phi(\mathbf{x})$ is continuous and injective.
\end{theorem}
\begin{proof}
We prove the theorem by three steps, 1. constructing a satisfying function $\Phi(x)$ in one dimensional cases ($d=1$); 2. constructing a satisfying function $\Phi(\mathbf{x})$ in multi-dimensional cases ($d>1$); 3. The number of the satisfying functions is infinite.

\noindent\textbf{1. One dimensional cases ($d=1$):}

In one dimensional case, the theorem can be easily proved by extending the following lemma from \cite{zaheer2017deep}.
\begin{lemma*}
Let $\mathcal{X}=\{(x_1,\cdots,x_M)\in [0,1]^M: x_1\leq x_2 \leq \cdots \leq x_M\}$. The sum-of-power mapping $E: \mathcal{X}\rightarrow \mathbb{R}^{M+1}$ defined by the coordinate functions 
\begin{equation}
    E(X)=\left [E_0(X),E_1(X),\cdots,E_M(X) \right]=\left[\sum_{x\in X}{1},\sum_{x\in X}{x},\cdots, \sum_{x\in X}{x^M} \right]
\end{equation}
is injective.
\end{lemma*}
In \cite{zaheer2017deep}, this lemma is proved based on the famous Newton-Girard formulae, where the domain $\mathcal{X}$ can be extended to $\mathcal{X}=\{X|X\subset \mathbb{R}, |X|\leq M, M\in \mathbb{N}\}$ with the same proof process.
Because $E_0(X)=\sum_{x\in X} {1}=|X|$ is the number of elements in $X$, $E_0(X_1)=E_0(X_2)$ implies equal set size between the two sets, it can be easily extended to $\mathcal{X}=\{X|X\subset \mathbb{R}, |X|\leq M, M\in \mathbb{N}\}$.

Since $E(X)=\left[\sum_{x\in X}{1},\sum_{x\in X}{x},\cdots, \sum_{x\in X}{x^M} \right] = \sum_{x\in X}{[1,x,\cdots,x^M]}$. Let $\Phi(x)=[1,x,\cdots,x^M]$, obviously $\Phi(x)$ is continuous, thus, we get one $\Phi(X)$ such that $f(X)=\sum_{x\in X} \Phi(x)$ is injective and continuous. 

\noindent\textbf{2. Multi-dimensional cases ($d>1$):}

Consider the following function $\Phi:\mathbb{R}^d\rightarrow \mathbb{R}^{(dm+1)}$, 
\begin{equation}\label{eq:phi}
\Phi(\mathbf{x})=\left[ 
\begin{aligned} 
    1, & \mathbf{x}[1], & \mathbf{x}[1]^2,  & \cdots, & \mathbf{x}[1]^{(M-1)}, & \mathbf{x}[1]^{M},   \\
    \mathbf{x}[2], & \mathbf{x}[1]\mathbf{x}[2], & \mathbf{x}[1]^2\mathbf{x}[2], & \cdots, & \mathbf{x}[1]^{M-1}\mathbf{x}[2],  \\
    \vdots  & \vdots & \vdots  & \ddots & \vdots \\
    \mathbf{x}[d], & \mathbf{x}[1]\mathbf{x}[d], & \mathbf{x}[1]^2\mathbf{x}[d], & \cdots, & \mathbf{x}[1]^{M-1}\mathbf{x}[d] 
    \end{aligned}
    \right] 
\end{equation}
where $\mathbf{x}[i]$ is the $i$th entry of vector $\mathbf{x}$. For a ($dm+1$)-dimensional vector $V$ from the image domain of $\mathcal{X}$ through $f(X)=\sum_{\mathbf{x}\in X} \Phi(\mathbf{x})$, we will identify the number of the preimages of $V$. 

Let $X$ is a preimage of $V$, we have the following equation which is exactly a equation group with $dM+1$ equations.
\begin{equation}\label{eq:sumphi}
V=
\left[ 
\begin{matrix} 
    \sum_{\mathbf{x} \in X}1, & \sum_{\mathbf{x} \in X}\mathbf{x}[1], & \sum_{\mathbf{x} \in X}\mathbf{x}[1]^2,  & \cdots, & \sum_{\mathbf{x} \in X}\mathbf{x}[1]^{(M-1)}, & \sum_{\mathbf{x} \in X}\mathbf{x}[1]^{M},   \\
    \sum_{\mathbf{x} \in X}\mathbf{x}[2], & \sum_{\mathbf{x} \in X}\mathbf{x}[1]\mathbf{x}[2], & \sum_{\mathbf{x} \in X}\mathbf{x}[1]^2\mathbf{x}[2], & \cdots, & \sum_{\mathbf{x} \in X}\mathbf{x}[1]^{M-1}\mathbf{x}[2],  \\
    \vdots  & \vdots & \vdots  & \ddots & \vdots \\
    \sum_{\mathbf{x} \in X}\mathbf{x}[d], & \sum_{\mathbf{x} \in X}\mathbf{x}[1]\mathbf{x}[d], & \sum_{\mathbf{x} \in X}\mathbf{x}[1]^2\mathbf{x}[d], & \cdots, & \sum_{\mathbf{x} \in X}\mathbf{x}[1]^{M-1}\mathbf{x}[d] 
    \end{matrix}
    \right]  
\end{equation}

Note that the first row of Eq. \ref{eq:sumphi} is exactly the sum-of-power mapping we considered in one dimensional cases, thus we can identify a unique set of the first entry of elements in $X$, and the number of elements in $X$ is also determined. Let $X$ have $M$ elements and $X=\{\mathbf{x_1},\mathbf{x_2},\cdots,\mathbf{x_M}\}$, the set $\{\mathbf{x_1}[1],\mathbf{x_2}[1],\cdots,\mathbf{x_M}[1]\}$ is uniquely defined.

Consider the second row of  Eq. \ref{eq:sumphi}, we can rewrite the equations in the second row as linear matrix equation as Eq. \ref{eq:eqvdmd}
\begin{equation}\label{eq:eqvdmd}
\left[ 
\begin{matrix} 
    1 & 1  & \cdots & 1   \\
    \mathbf{x_1}[1] & \mathbf{x_2}[1] &  \cdots & \mathbf{x_M}[1]  \\
    \mathbf{x_1}[1]^2 & \mathbf{x_2}[1]^2 &  \cdots & \mathbf{x_M}[1]^2  \\
    \vdots  & \vdots & \ddots  & \vdots  \\
    \mathbf{x_1}[1]^{(M-1)} & \mathbf{x_2}[1]^{(M-1)} &  \cdots & \mathbf{x_M}[1]^{(M-1)}  \\
\end{matrix}
\right] \times 
\left[
\begin{matrix}
    \mathbf{x_1}[2]  \\
          \mathbf{x_2}[2]  \\
          \vdots  \\
          \mathbf{x_M}[2]  \\
    
\end{matrix}
\right]
= V[2,:]^T
\end{equation} 

Note that the coefficient matrix in left side of Eq. \ref{eq:eqvdmd} is a Vandermonde matrix, if the $\mathbf{x_1}[1],\cdots, \mathbf{x_M}[1]$ are all distinct, the coefficient matrix is invertible \cite{macon1958inverses}, Eq. \ref{eq:eqvdmd} has a unique solution for $\mathbf{x_1}[2],\cdots, \mathbf{x_M}[2]$ corresponding to $\mathbf{x_1}[1],\cdots, \mathbf{x_M}[1]$. Similarly, by the $i$th ($2<i<d$) row of Eq. \ref{eq:sumphi}, $\mathbf{x_1}[i],\cdots, \mathbf{x_M}[i]$ can be uniquely identified. 

In the other case, if the $\mathbf{x_1}[1],\cdots, \mathbf{x_M}[1]$ that solved from the first row of Eq. \ref{eq:sumphi} are not all distinct, Eq. \ref{eq:eqvdmd} has infinitely many solutions. $\Phi(\mathbf{x})$ defined by Eq. \ref{eq:phi} is not sufficient to make $f(X)=\sum_{\mathbf{x}\in X} \Phi(\mathbf{x})$ injective. We need some more dimensions appended in $\Phi(\mathbf{x})$. 

Let $\mathbf{x_1}[1]= \mathbf{x_2}[1] = \cdots =\mathbf{x_k}[1]$, then by combining the items, Eq. \ref{eq:eqvdmd} is shrinked to
\begin{equation}\label{eq:shrinkvdmd}
\left[ 
\begin{matrix} 
    1 & 1  & \cdots & 1   \\
    \mathbf{x_k}[1] & \mathbf{x_{k+1}}[1] &  \cdots & \mathbf{x_M}[1]  \\
    \mathbf{x_k}[1]^2 & \mathbf{x_{k+1}}[1]^2 &  \cdots & \mathbf{x_M}[1]^2  \\
    \vdots  & \vdots & \ddots  & \vdots  \\
    \mathbf{x_k}[1]^{(M-1)} & \mathbf{x_{k+1}}[1]^{(M-1)} &  \cdots & \mathbf{x_M}[1]^{(M-1)}  \\
\end{matrix}
\right] \times 
\left[
\begin{matrix}
    \sum_{i=1\cdots k}\mathbf{x_i}[2]  \\
          \mathbf{x_{k+1}}[2]  \\
          \vdots  \\
          \mathbf{x_M}[2]  \\
    
\end{matrix}
\right]
= V[2,:]^T
\end{equation} 
By solving Eq. \ref{eq:shrinkvdmd}, we have a unique sum $\sum_{i=1\cdots k}\mathbf{x_i}[2]$. To identify a unique set of $\{\mathbf{x_1}[2], \mathbf{x_2}[2], \cdots, \mathbf{x_k}[2]\}$, we can define a unique $\sum_{i=1\cdots k}\mathbf{x_i}[2]^2, \sum_{i=1\cdots k}\mathbf{x_i}[2]^3, \cdots, \sum_{i=1\cdots k}\mathbf{x_i}[2]^k$, we can add items $\mathbf{x}[2]^2, \mathbf{x}[1]\mathbf{x}[2]^2, \mathbf{x}[1]^2\mathbf{x}[2]^2, \cdots, \mathbf{x}[1]^{M-1}\mathbf{x}[2]^2$ to $\Phi(\mathbf{x})$ to uniquely identify $\sum_{i=1\cdots k}\mathbf{x_i}[2]^2$. Similarly, add items $\mathbf{x}[2]^k, \mathbf{x}[1]\mathbf{x}[2]^k, \mathbf{x}[1]^2\mathbf{x}[2]^k, \cdots, \mathbf{x}[1]^{M-1}\mathbf{x}[2]^k$ to $\Phi(\mathbf{x})$ to uniquely identify $\sum_{i=1\cdots k}\mathbf{x_i}[2]^k$. Thus all $\mathbf{x_i}[2]$ are identified.

After the set $\{\mathbf{x_1}[2],\mathbf{x_2}[2],\cdots,\mathbf{x_M}[2]\}$ is uniquely defined, by adding $\mathbf{x}[2]^i\mathbf{x}[j] (i=0,\cdots M-1, j=3,\cdots d)$ to $\Phi(\mathbf{x})$, we can use $\mathbf{x_i}[2]$ to construct a Vandermonde matrix to solve $\mathbf{x_i}[3], \cdots, \mathbf{x_i}[d], (i=1,\cdots,M)$. If $\mathbf{x_i}[1:2]$ are not all distinct, we can identify $\mathbf{x_i}[3]$ similarly by adding $\mathbf{x}[1]^i\mathbf{x}[3]^j$ to $\Phi(\mathbf{x})$. By this way, the set $X$ can be uniquely identified.

In our construction of $\Phi(\mathbf{x})$, all the functions are continuous, thus, there exists a continuous function $\Phi(\mathbf{x})$ such that $f(X)=\sum_{\mathbf{x}\in X} \Phi(\mathbf{x})$ is injective and continuous.

\noindent\textbf{3. The number of the satisfying function is infinity:}

To prove the number of this kind functions is infinity, we construct a continuous injective function $g:\mathbb{R^d}\rightarrow \mathbb{R^d}$, we will show that if we have a $\phi(\mathbf{x})$ satisfying the condition $f(X)=\sum_{\mathbf{x}\in X}{\phi(\mathbf{x})}$ is continuous and injective, then $\phi(g(\mathbf{x}))$ also satisfy the condition $f(X)=\sum_{\mathbf{x}\in X}{\phi(g(\mathbf{x}))}$ is continuous and injective.

We define a function $h:\mathcal{X}\rightarrow \mathcal{X}$, $h(X)=\{g(\mathbf{x})| \mathbf{x}\in X\}$, since $g(\mathbf{x})$ is injective, $h(X)$ is also injective. If we have a function $\phi(\mathbf{x})$ such that $f(X)=\sum_{\mathbf{x}\in X} \phi(\mathbf{x})$ is injective, $f(h(X))$ is injective. 
\begin{equation}
f(h(X))= \sum_{\mathbf{x}\in h(X)}{\phi(\mathbf{x})}=\sum_{\mathbf{x}\in\{g(\mathbf{x})|\mathbf{x}\in X\}} {\phi(\mathbf{x})}  
=\sum_{\mathbf{x}\in X} {\phi(g(\mathbf{x}))}  
\end{equation}
Because $\phi(\mathbf{x})$ and $g(\mathbf{x})$ are both continuous, $\phi(g(\mathbf{x}))$  is continuous, thus we find another function $\phi(g(\mathbf{x}))$ such that $\sum_{\mathbf{x}\in X} {\phi(g(\mathbf{x}))}$ is injective. Because we can have an infinite number of such continuous injective functions $g:\mathbb{R}^d\rightarrow \mathbb{R}^d$ (e.g., $g(\mathbf{x})=k\mathbf{x}, k\in \mathbb{R}$), we have a infinite number of such functions $\Phi(\mathbf{x})=\phi(g(\mathbf{x}))$ such that $f(X)=\sum_{\mathbf{x}\in X} \Phi(\mathbf{x})$ is injective.

\end{proof}

\begin{lemma} \label{lm:limitation}
Let $M\in \mathbb{N}$ and $\mathcal{X}=\{X | X \subset \mathbb{R}^d, |X|=M \}$, then for any continuous function $\Phi : \mathbb{R^d}\rightarrow \mathbb{R}^{N}$, if $N < dM$, the set function $f:\mathcal{X}\rightarrow \mathbb{R}^{N}$ $f(X)=\sum_{\mathbf{x}\in X} \Phi(\mathbf{x})$ is not injective.
\end{lemma}
\begin{proof}
Suppose $f(X)$ is injective. Because $\Phi(\mathbf{x})$ is continuous, $f(X)$ is a finite sum of continuous function, it is also continuous, thus, $f(X)$ is continuous and injective.

All sets in $\mathcal{X}$ have $M$ elements from $\mathbb{R}^d$. In one dimensional cases ($d=1$), $\mathcal{X}$ has a bijection to  $\mathcal{S}=\{X=(x_1,\cdots,x_M)| X\in \mathbb{R}^M, x_1\leq x_2\leq\cdots\leq x_M\}$.

In multi-dimensional cases ($d>1$), we can construct a bijection from $\mathcal{X}$ to $\mathcal{S}=\{X=(\mathbf{x_1},\cdots,\mathbf{x_M})| X\in \mathbb{R}^{dM}, \mathbf{x_1}[1]\leq \mathbf{x_2}[1]\leq\cdots\leq \mathbf{x_M}[1], \text{if } \mathbf{x_i}[1:k]=\mathbf{x_{i+1}}[1:k], \mathbf{x_i}[k+1]\leq\mathbf{x_{i+1}}[k+1], i=1,\cdots M-1, k=1,\cdots,d-1\}$. For $X \in \mathcal{X}$, let $X=\{\mathbf{x_1},\cdots,\mathbf{x_M}\}$, we can sort elements in $X$ by the first entry, for the elements whose first entries are equal, sort them by the second entry, so repeatedly in this way, we get a final ordered sequence of the vectors, which is unique in $\mathcal{S}$.   

Note that $\mathcal{S}$ is a convex open subset of $\mathbb{R}^{dM}$, and is therefore homeomorphic to $\mathbb{R}^{dM}$. Since $N<dM$, no continuous injection exists from $\mathbb{R}^{dM}$ to $\mathbb{R}^N$. Thus no continuous injective function exist from $\mathcal{X}$ to $\mathbb{R}^N$. Hence we have reached a contradiction.
\end{proof}

\section{Dataset Details}
Table \ref{tab:dataset} is the dataset information. All datasets are available from \cite{KKMMN2016}.

\begin{table*}[ht]
\small
  \centering
  \caption{Dataset information. \#G=number of graphs. \#C=number of classes. AvgN=average number of nodes in one graph. AvgE=average number of edges in one graph. Dim=node attribute dimension. MaxNeighb is the max 1-hop neighbors in all the nodes. }
    \begin{tabular}{l|c|c|c|c|c|c|c|l}
    \toprule
          & \#G & \#C & AvgN & AvgE  & MaxNeighb & Dim & Type & Source \\
\midrule       
   MUTAG &188 &	2	&17.93&	19.79 & 4 & -- & bioinformatics & \cite{debnath1991structure, kriege2012subgraph}\\
   PTC &344	&2	&14.29&	14.69 & 4 & --  & bioinformatics &  \cite{helma2001predictive, kriege2012subgraph}\\
   NCI1 &4110&	2	&29.87&	32.30 & 4 & --  & bioinformatics & \cite{wale2008comparison, shervashidze2011weisfeiler}\\
   PROTEINS &1113&	2	&39.06&	72.82 & 25 & --  & bioinformatics & \cite{borgwardt2005protein, dobson2003distinguishing}\\
   COLLAB &5000&	3	&74.49&	2457.78 & 491 & -- & social networks  & \cite{yanardag2015deep}\\
    IMDB-B & 1000&	2	&19.77&	96.53 & 135 &-- & social networks  & \cite{yanardag2015deep}\\
    IMDB-M & 1500&	3	&13.00&	65.94 &  88  & -- & social networks & \cite{yanardag2015deep}\\
    RDT-B & 2000 &	2	&429.63&	497.75 & 3062 & -- & social networks & \cite{yanardag2015deep}\\
    \midrule
    ENZYMES & 600&	6	&32.63&	62.14& 9 & 18  & bioinformatics & \cite{borgwardt2005protein, schomburg2004brenda}\\
    FRANKENSTEIN & 4337&	2&	16.90&	17.88& 4 & 780  & bioinformatics & \cite{orsini2015graph}\\
    PROTEINS-att &1113&	2	&39.06&	72.82 & 25 & 1  & bioinformatics & \cite{borgwardt2005protein, dobson2003distinguishing}\\
    SYNTHETICnew & 300	&2&	100.00&	196.25 & 9 & 1 & synthetic & \cite{feragen2013scalable}\\
    Synthie &400&	4	&95.00&	172.93 & 20 & 15 & synthetic & \cite{morris2016faster}\\
    \bottomrule
    \end{tabular}%
  \label{tab:dataset}%
\end{table*}%

\section{Results}
\begin{figure}[h]
  \centering
  \subfloat[PROTEINS]{\includegraphics[width=.33\textwidth,page=1]{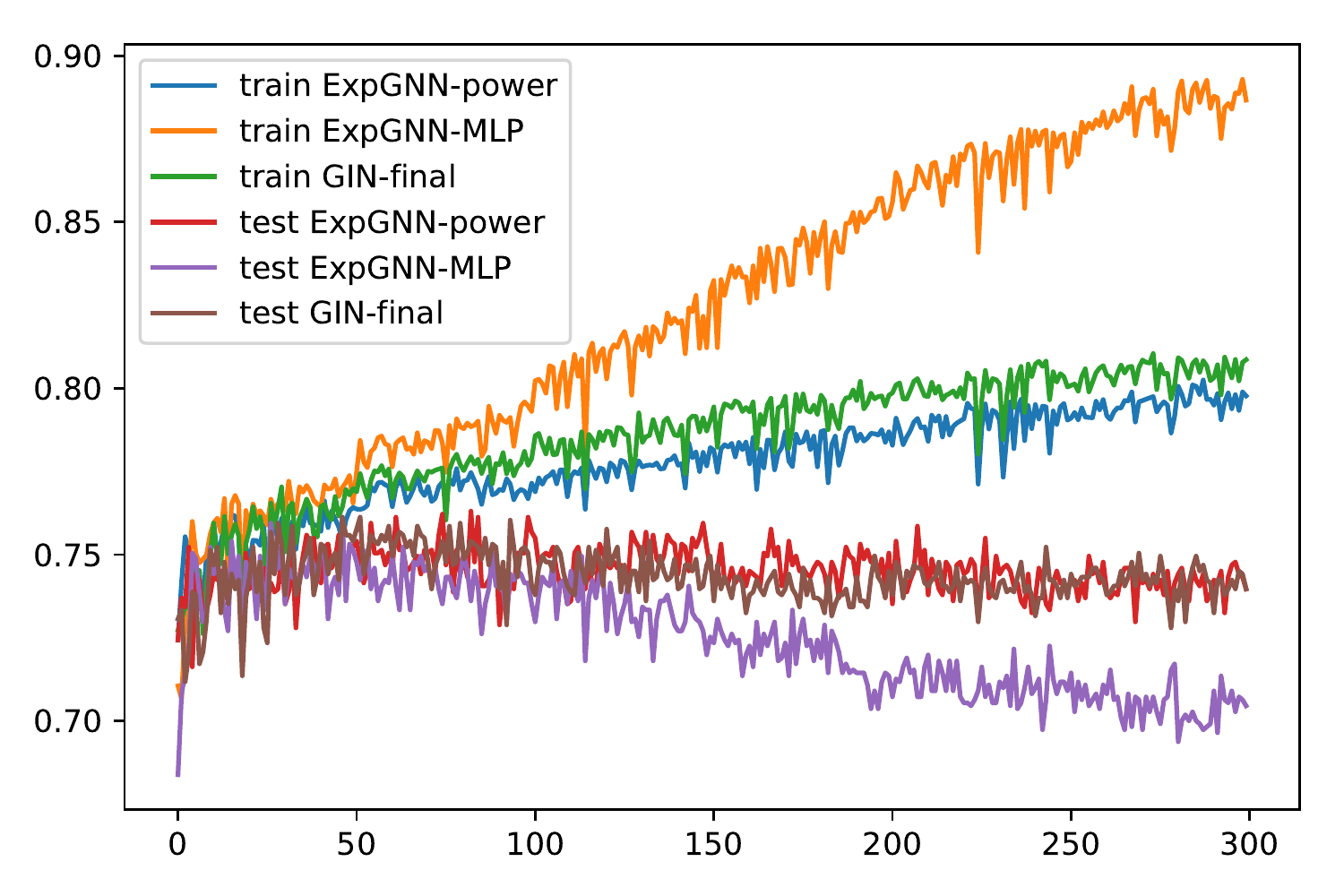}}
  \subfloat[IMDBBINARY]{\includegraphics[width=.33\textwidth,page=1]{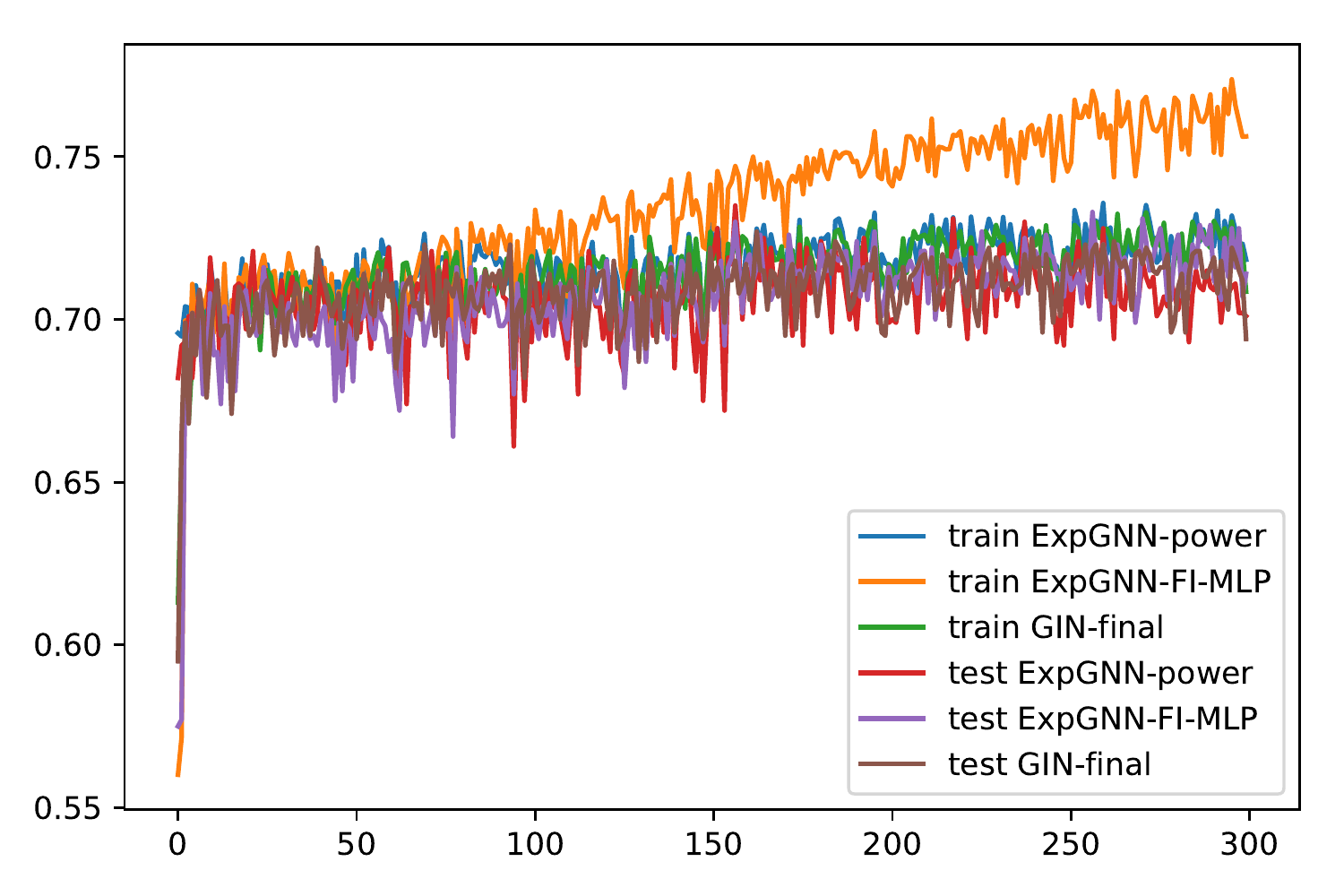}}
  \subfloat[IMDBMULTI]{\includegraphics[width=.33\textwidth,page=1]{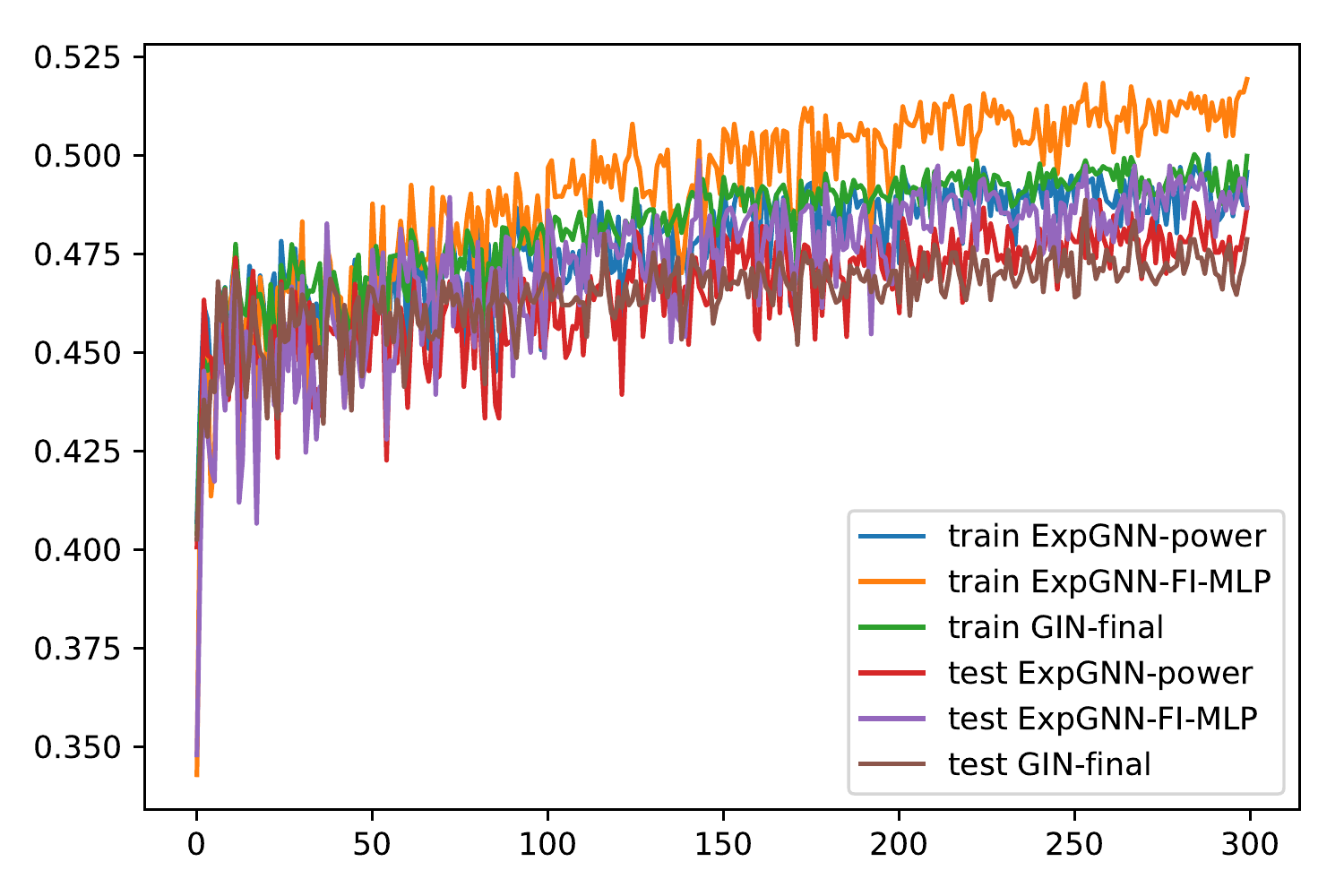}}

  \subfloat[ENZYMES]{\includegraphics[width=.33\textwidth,page=1]{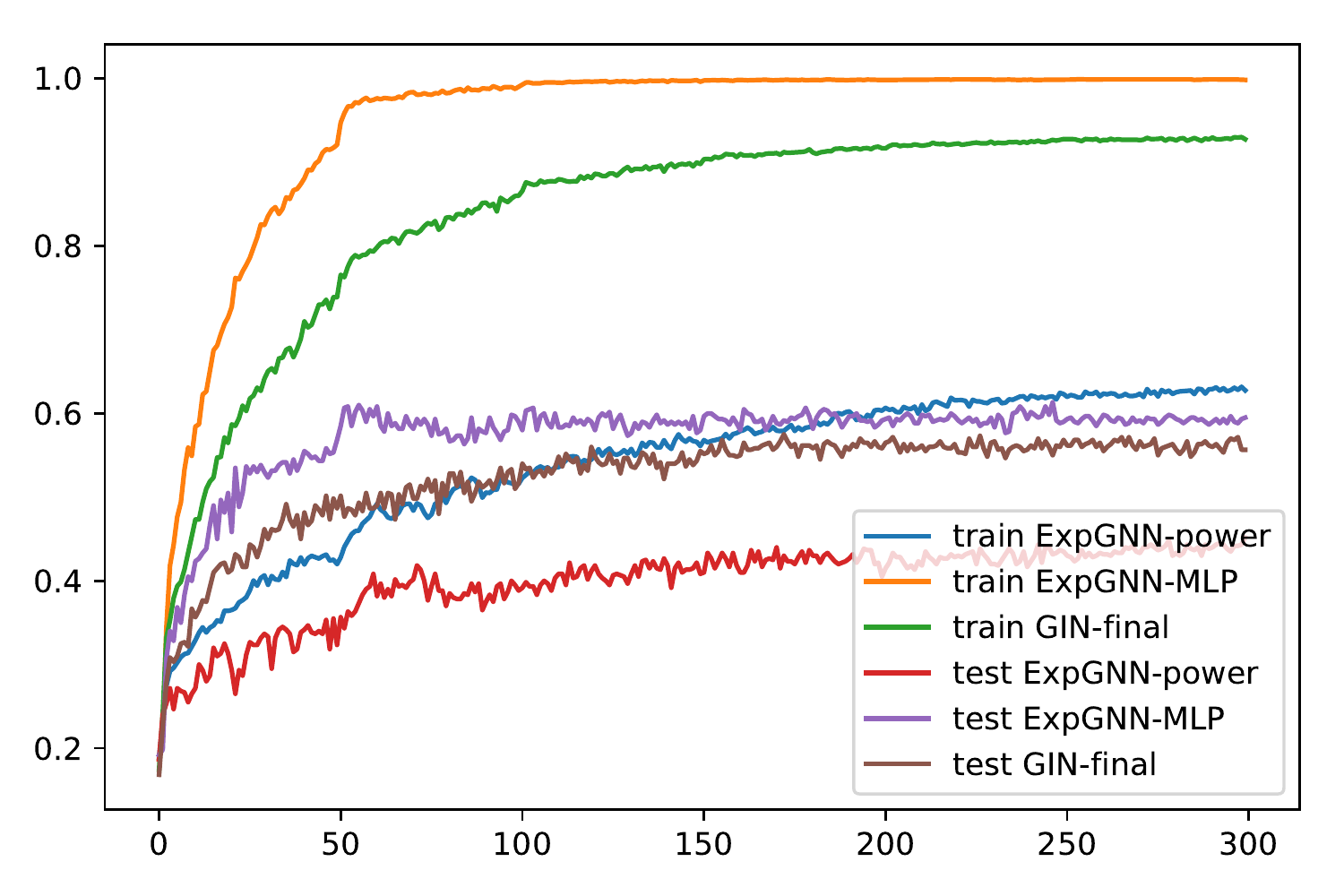}}
  \subfloat[SYNTHETICnew]{\includegraphics[width=.33\textwidth,page=1]{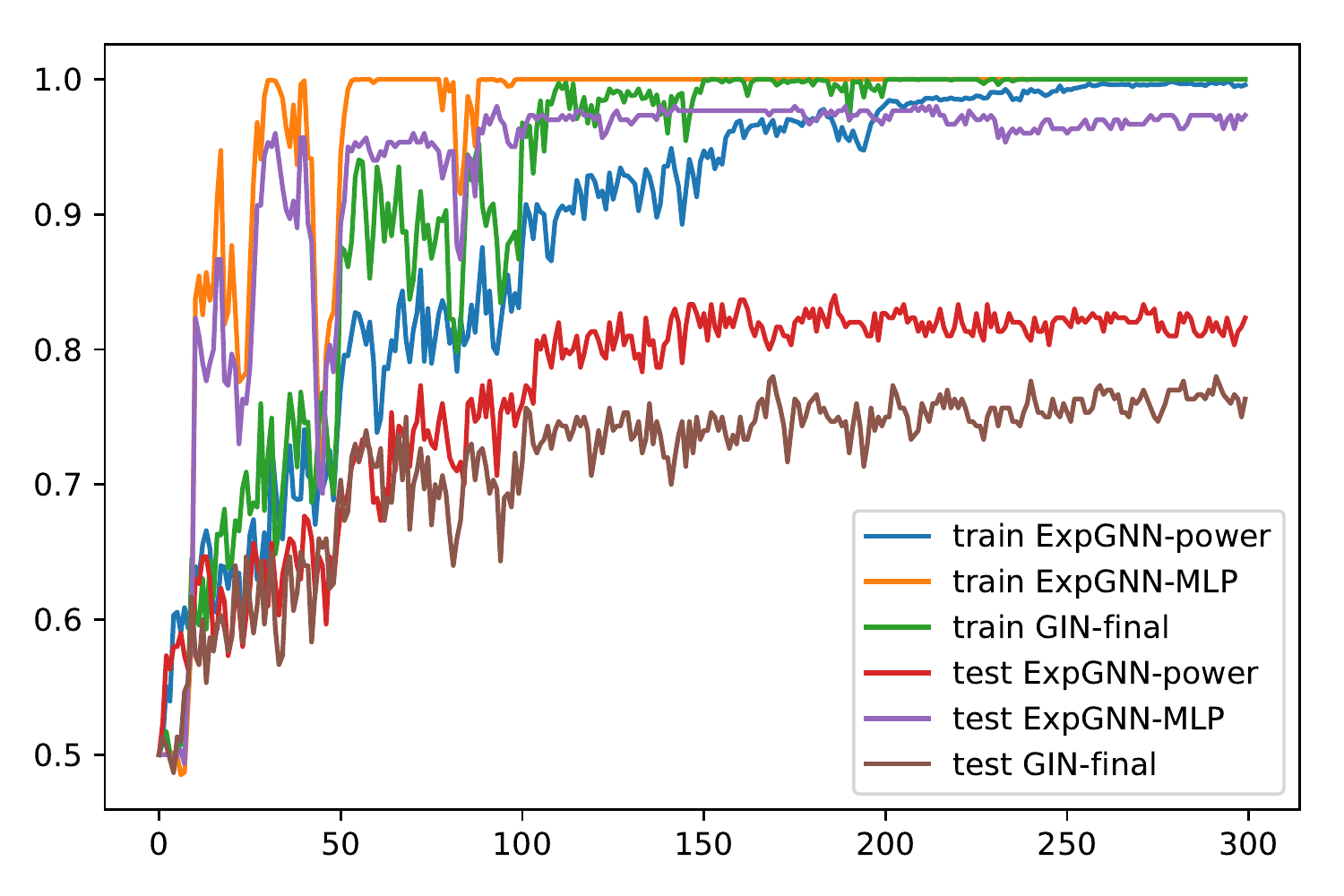}}
  \subfloat[Synthie]{\includegraphics[width=.33\textwidth,page=1]{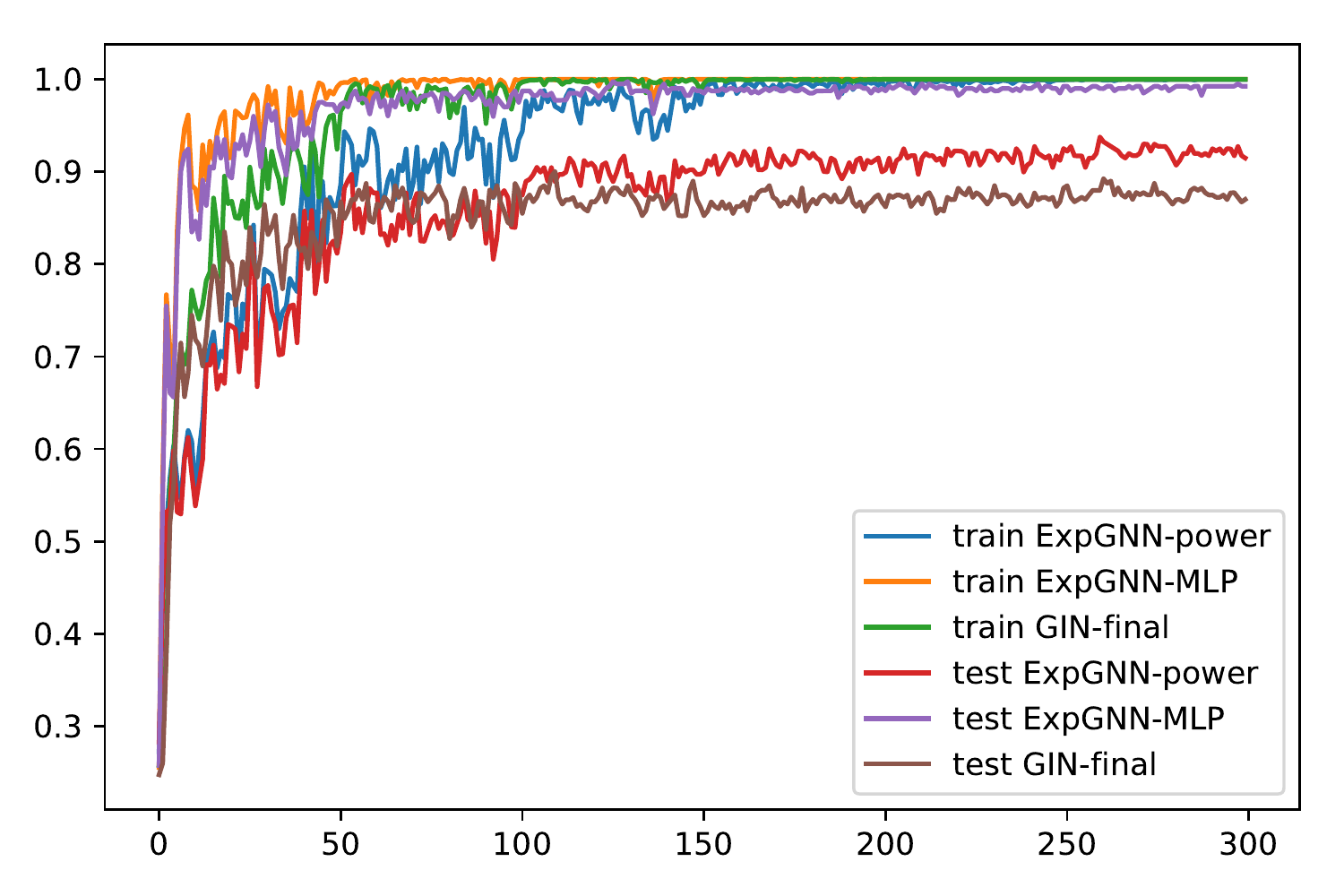}}
  \caption{The accuracy curves on training set and test set in the training process.}
  \label{fig:traincurve}
\end{figure}

The accuracy curves on training set and test set in the training process for different datasets are shown in Figure \ref{fig:traincurve}.

\begin{figure}[h]
  \centering
  \subfloat[ExpGNN-MLP]{\includegraphics[width=.33\textwidth,page=1]{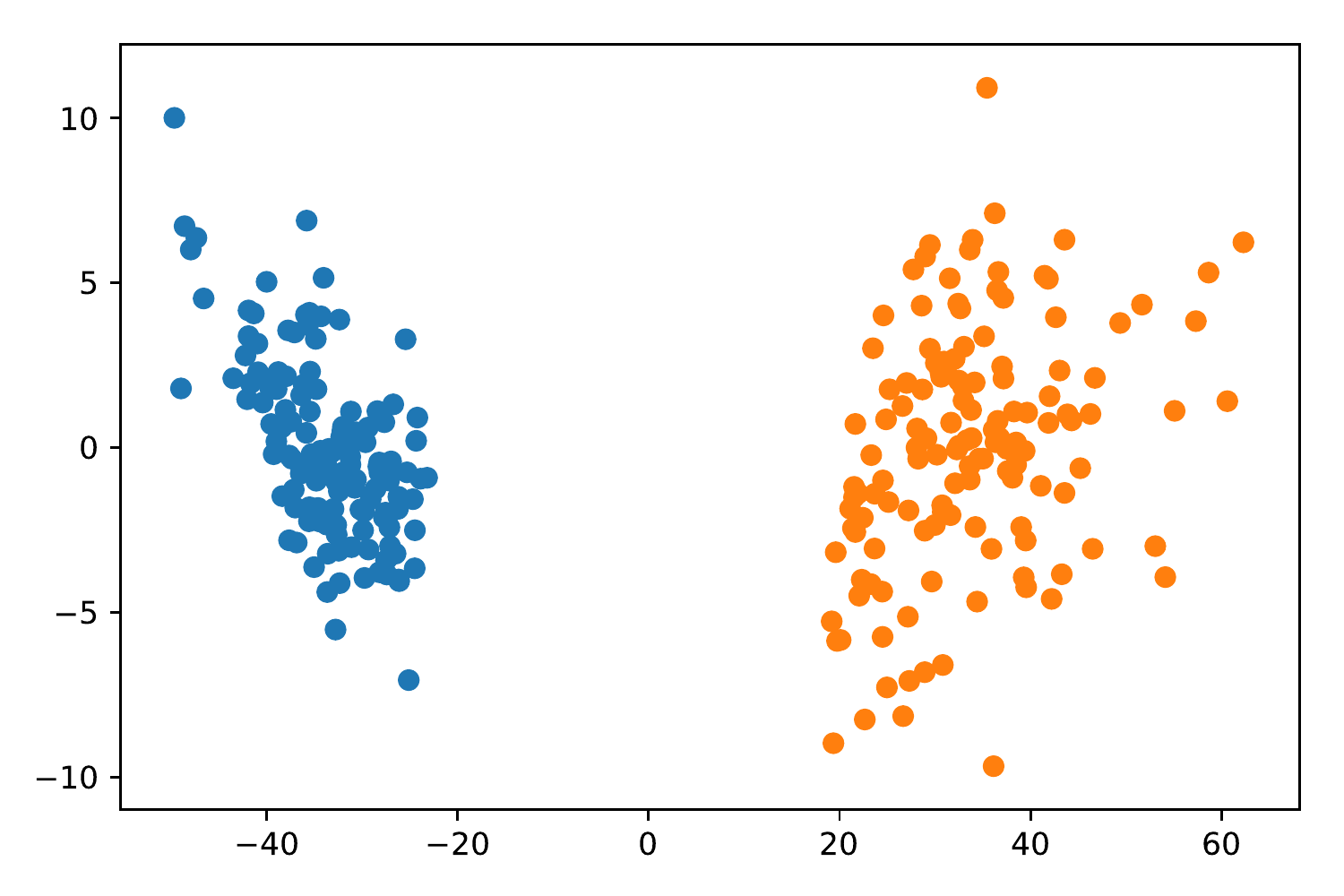}}
  \subfloat[ExpGNN-power]{\includegraphics[width=.33\textwidth,page=1]{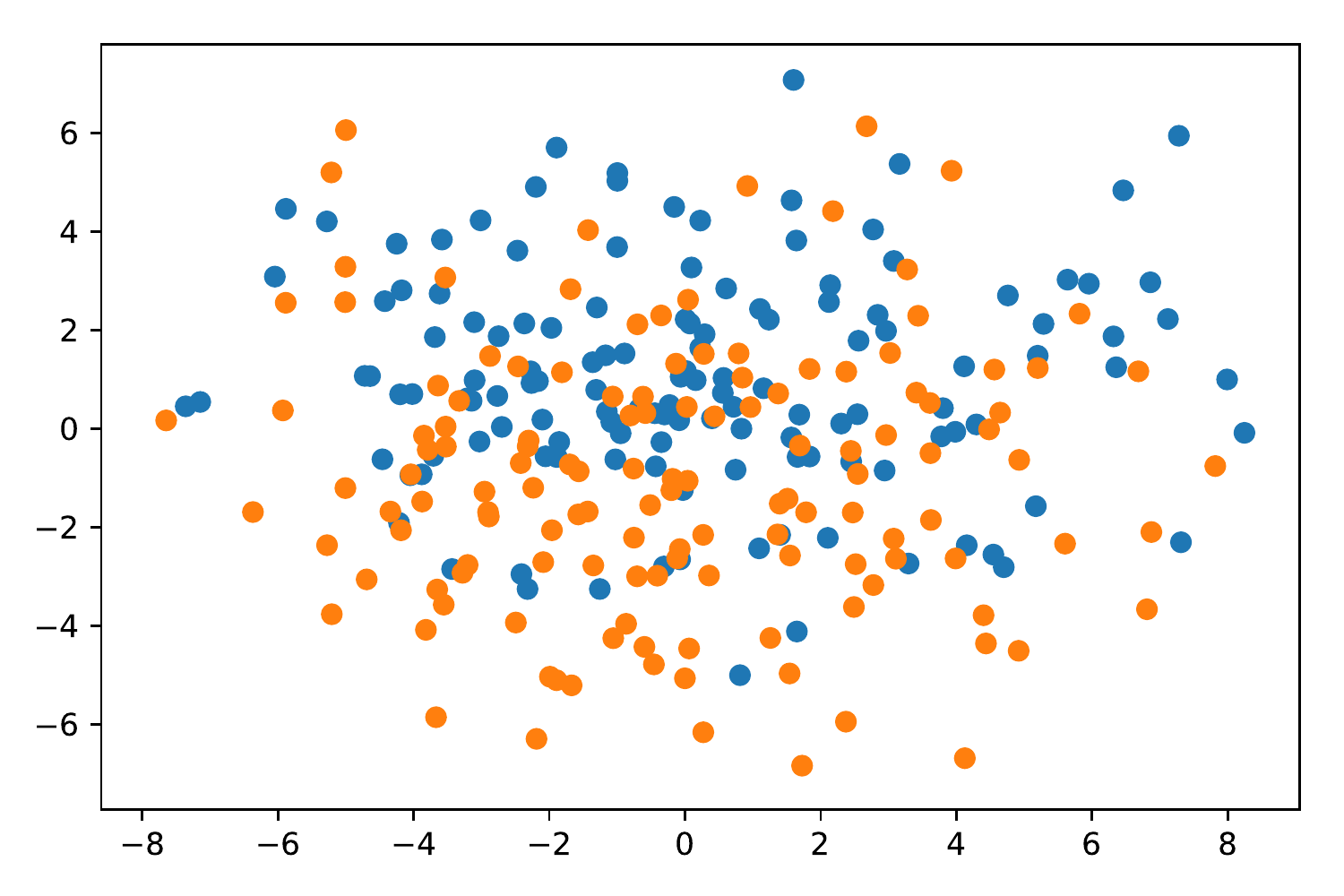}}
  \subfloat[GIN-final]{\includegraphics[width=.33\textwidth,page=1]{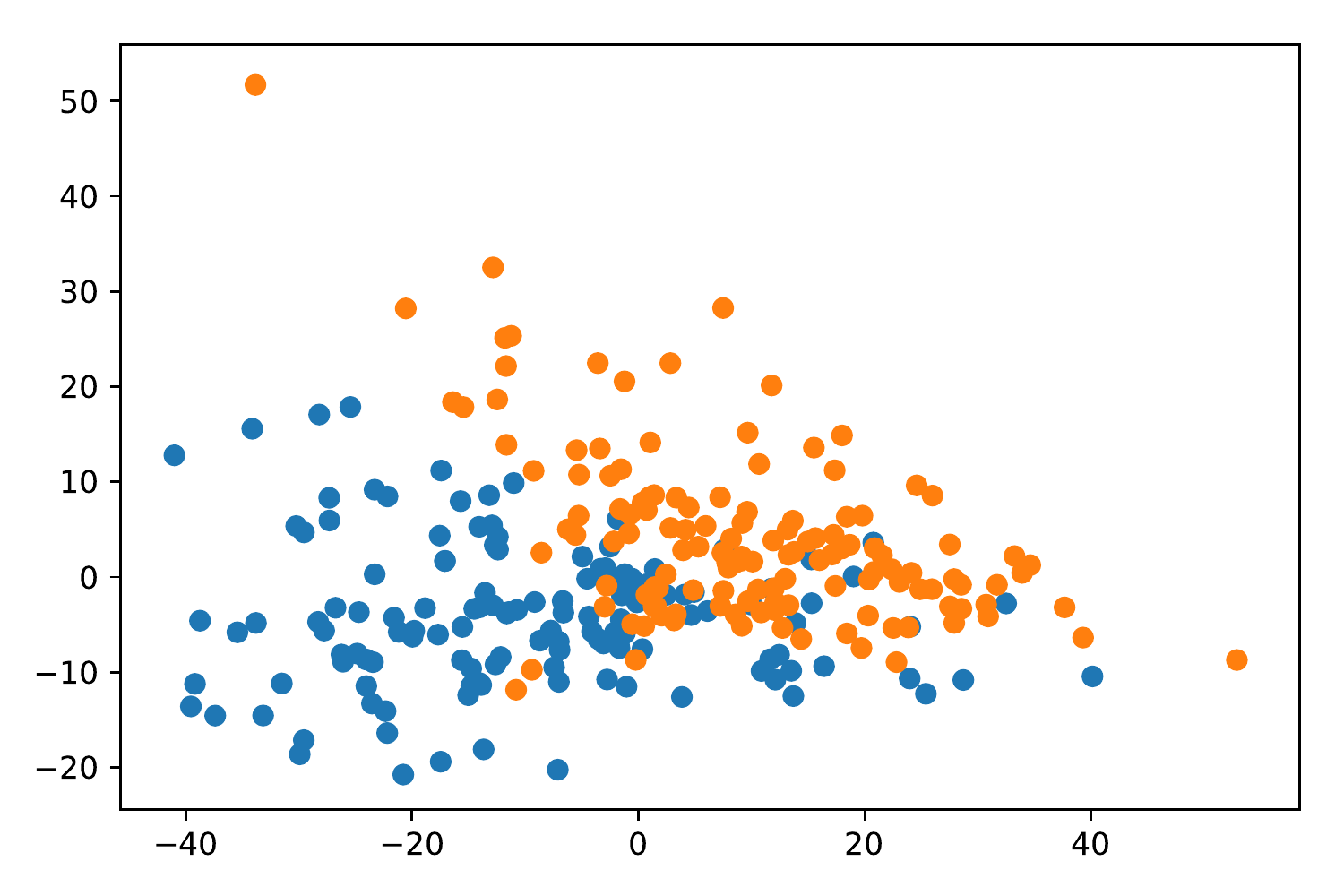}}
  \caption{PCA visualization of the output embeddings on training data of Synthie dataset.}
  \label{fig:pcavisual}
\end{figure}

\begin{figure}[h]
  \centering
  \subfloat[ExpGNN-MLP]{\includegraphics[width=.33\textwidth,page=1]{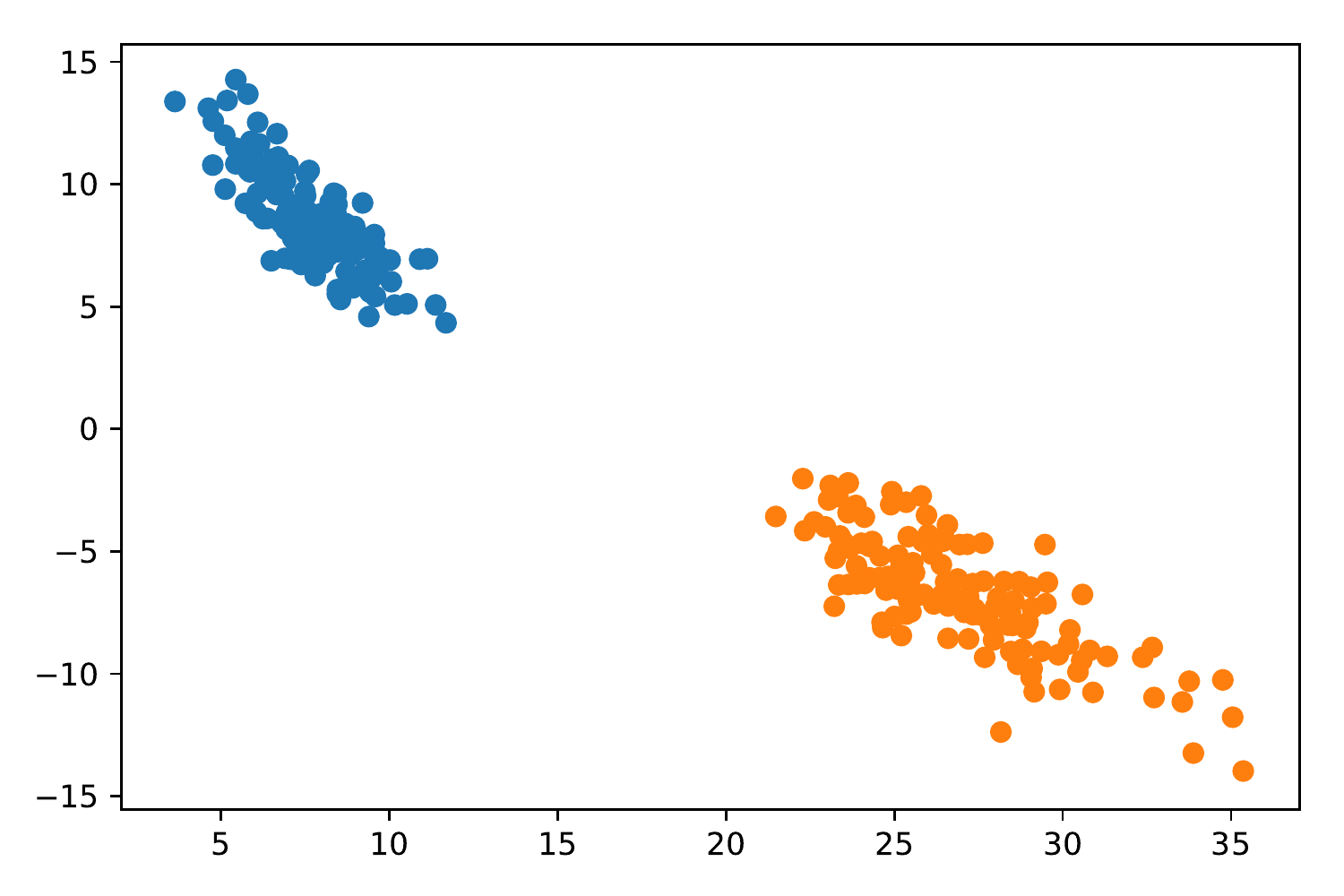}}
  \subfloat[ExpGNN-power]{\includegraphics[width=.33\textwidth,page=1]{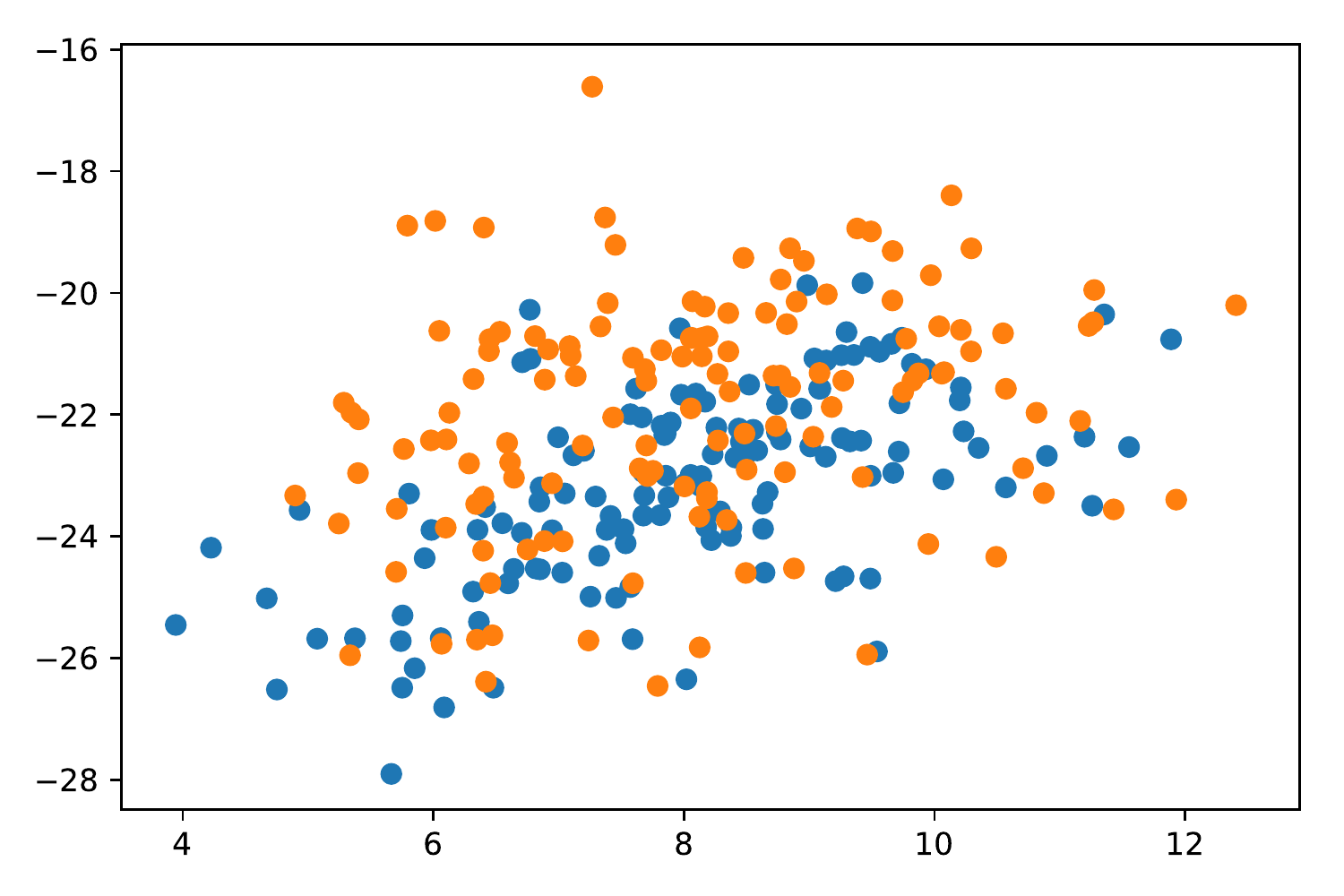}}
  \subfloat[GIN-final]{\includegraphics[width=.33\textwidth,page=1]{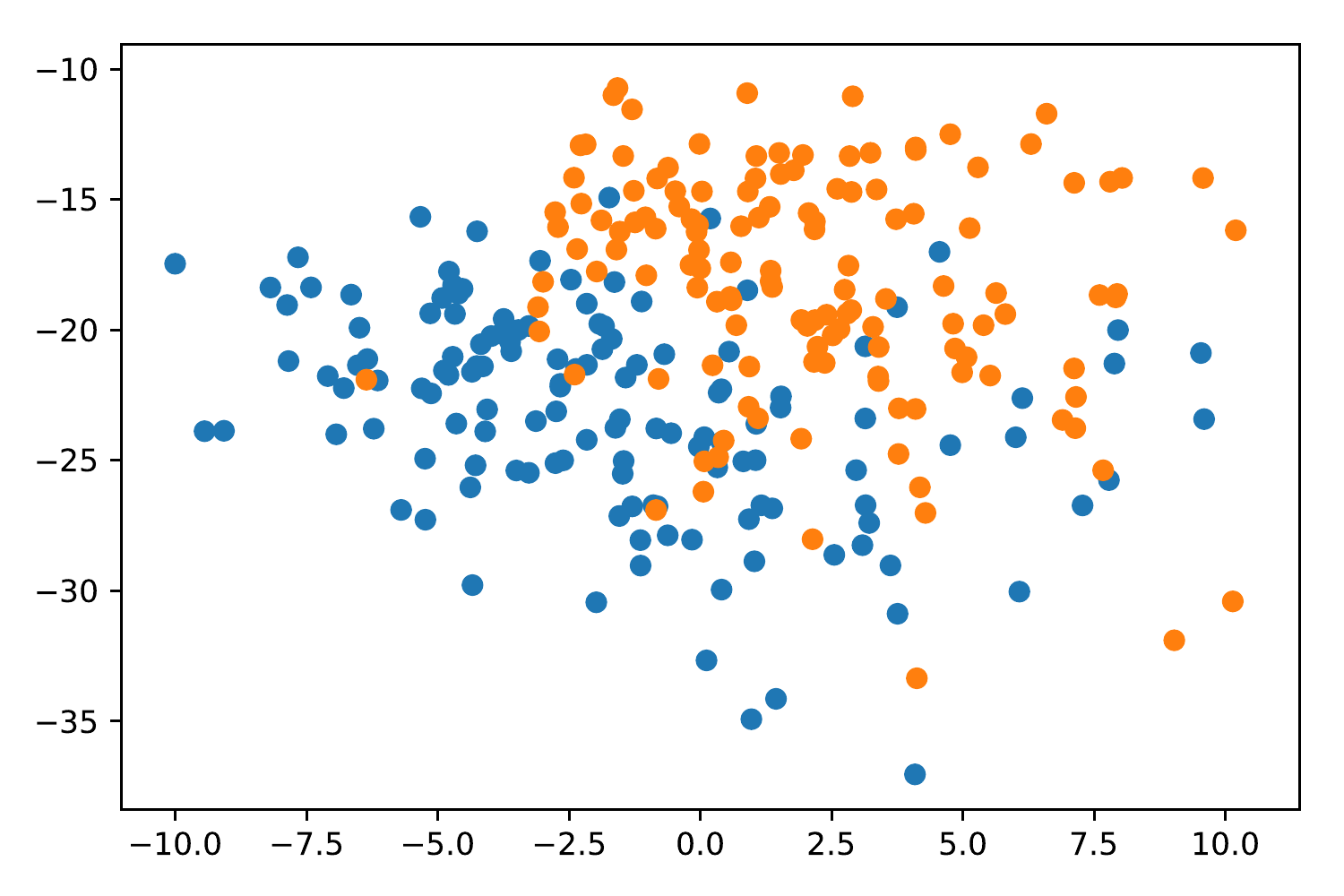}}
  \caption{visualization of the output embeddings with 2 random dimensions on training data of Synthie dataset.}
  \label{fig:randvisual}
\end{figure}

The final layer GNN output embedding in SYNTHETICnew dataset is visualized with PCA (Figure \ref{fig:pcavisual}) and two random dims (Figure \ref{fig:randvisual}). 

\bibliographystyle{plain}
\bibliography{references}